\documentclass[lettersize,journal]{IEEEtran}
\usepackage{amsmath,amsfonts}
\usepackage{algorithmic}
\usepackage{algorithm}
\usepackage{array}
\usepackage[caption=false,font=normalsize,labelfont=sc,textfont=sc]{subfig}
\usepackage{textcomp}
\usepackage{stfloats}
\usepackage{url}
\usepackage{verbatim}
\usepackage{graphicx}
\usepackage{cite}
\usepackage{booktabs}
\usepackage[pagebackref=false,breaklinks=true,colorlinks=true,bookmarks=false,citecolor=blue,linkcolor=blue,urlcolor=blue]{hyperref}
\usepackage[capitalize]{cleveref}
\crefname{section}{Sect.}{Secs.}
\crefname{table}{Tab.}{Tabs.}
\usepackage{bm}
\usepackage{multirow}
\usepackage{adjustbox}
\usepackage{pifont}
\usepackage{color, colortbl}
\definecolor{Gray}{gray}{0.9}

\newcommand{\cmark}{\ding{51}}
\newcommand{\xmark}{\ding{55}}
\newcommand{\etal}{\textit{et al.}}
\newcommand{\etc}{\textit{etc}}
\newcommand{\eg}{\textit{e.g.}}
\newcommand{\ieno}{\textit{i.e.}}
\newcommand{\et}[2]{${#1}^{\pm{#2}}$}

\begin{document}

\title{Inter-X++: A Comprehensive Benchmark for Multimodal Human-Human Interaction Analysis}

\author{Liang~Xu, Chengqun~Yang, Zili~Lin, Xintao~Lv, Yichao~Yan, Xin~Jin, Zhibo~Chen, Xiaokang~Yang,~\IEEEmembership{Fellow,~IEEE} and Wenjun~Zeng,~\IEEEmembership{Fellow,~IEEE}
\IEEEcompsocitemizethanks{
\IEEEcompsocthanksitem Liang~Xu, Chengqun~Yang, Zili~Lin, Xintao~Lv, Yichao~Yan and Xiaokang~Yang are with Shanghai Jiao Tong University. E-mail: \{liangxu, ycq0191, linzili111666, lvxintao, yanyichao, xkyang\}@sjtu.edu.cn
\IEEEcompsocthanksitem Liang~Xu, Zili Lin, Xin~Jin and Wenjun~Zeng are with Eastern Institute of Technology, Ningbo. (E-mail: \{liangxu, zllin, jinxin, wzeng\}@eitech.edu.cn)
\IEEEcompsocthanksitem Zhibo~Chen is with University of Science and Technology of China. (E-mail: chenzhibo@ustc.edu.cn)
\IEEEcompsocthanksitem Corresponding author: Yichao~Yan.}
}

\markboth{Journal of \LaTeX\ Class Files,~Vol.~14, No.~8, August~2021}%
{Shell \MakeLowercase{\textit{et al.}}: A Sample Article Using IEEEtran.cls for IEEE Journals}


\maketitle

\begin{abstract}
The capability to perceive and synthesize human-human interactions (HHIs) is fundamental to developing intelligent digital human systems and understanding humans as social beings. 
However, existing datasets and modeling approaches are fundamentally constrained by low-fidelity kinematics, the omission of dexterous hand gestures and a severe lack of rich multimodal annotations. Furthermore, fragmented interaction representations and inconsistent evaluation protocols also impede fair and rigorous benchmarking. 
To systematically address these bottlenecks, we present Inter-X++, a comprehensive and large-scale benchmark designed to empower versatile HHI analysis. Captured via a novel hybrid motion capture system, Inter-X++ provides 11,388 high-fidelity interaction sequences and over 8.1M frames, featuring precise whole-body movements and detailed finger articulations. 
Meanwhile, we enrich the data foundation with multifaceted annotations, including hierarchical fine-grained textual descriptions, interaction categories, causal interaction orders, the relationship and personality of the subjects, as well as vertex-level contact maps and physically regularized constraints. 
Leveraging these elaborate annotations, we formulate a unified testing ground comprising four categories of downstream tasks that symmetrically span both generative and perceptive paradigms. 
To eliminate benchmarking ambiguities, we systematically standardize the interaction representations and evaluation protocols. 
Finally, we go beyond dataset construction to propose OpenHHI, a single and unified HHI representation and modeling framework that jointly optimizes interaction reconstruction and semantic understanding. 
Extensive experiments reveal that OpenHHI achieves state-of-the-art performance on both downstream generation and perception tasks. This definitively proves that our unified representation successfully bridges interaction understanding and generation simultaneously, solidifying Inter-X++ as a robust foundation for future HHI advancements.

\end{abstract}


\section{Introduction}

\begin{figure*}[t]
  \centering
   \includegraphics[width=1.0\linewidth]{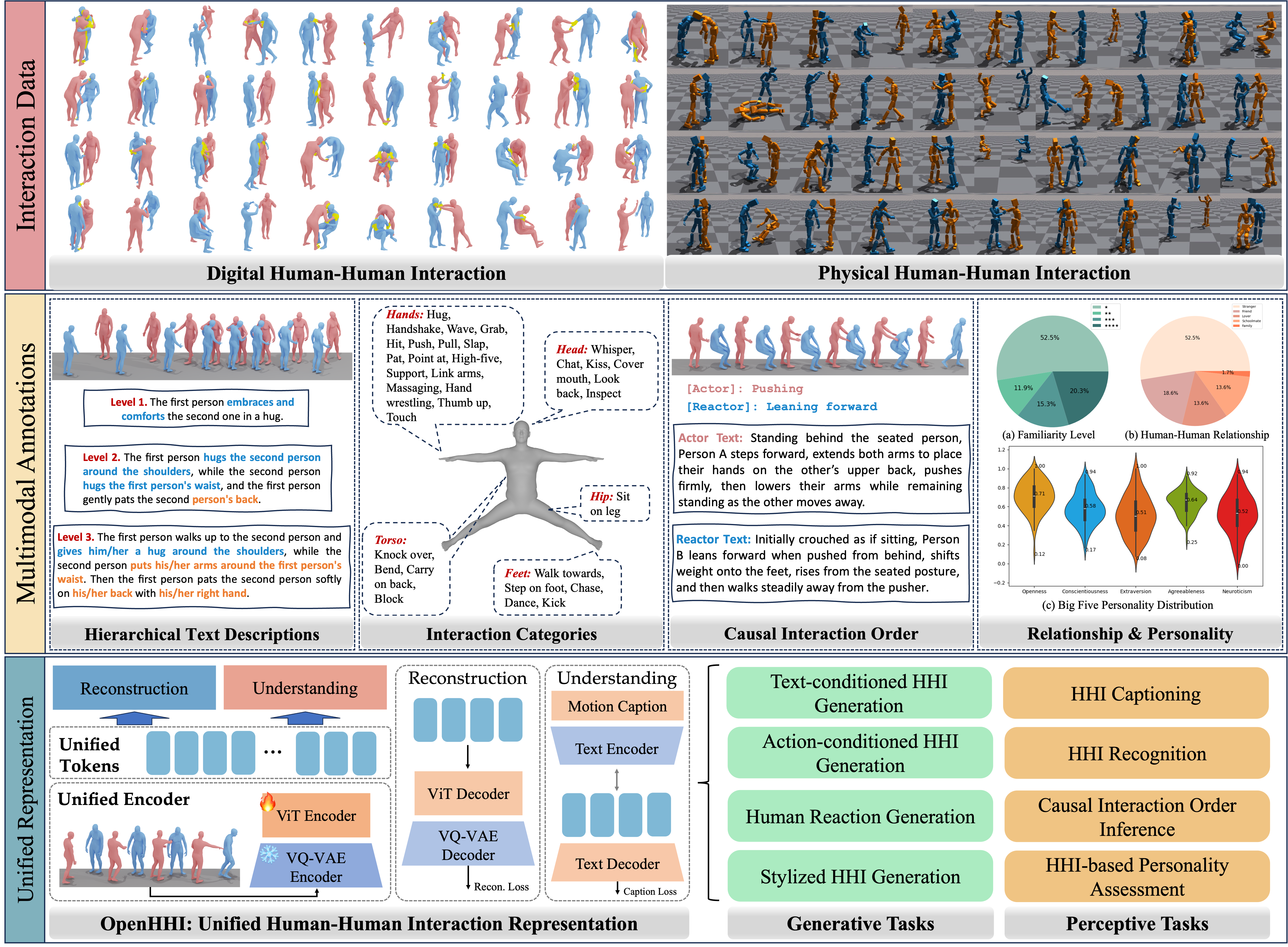}
   \caption{\textbf{Overview of Inter-X++.} We present Inter-X++ as a comprehensive benchmark for multimodal human-human interaction analysis structured across three dimensions of interaction data, multimodal annotations and unified representation. For interaction data, Inter-X++ contains high-fidelity MoCap results with vertex-level contact annotations while incorporating physical constraints. For multimodal annotations, we enrich Inter-X++ with hierarchical text descriptions, interaction categories, causal interaction order, relationship and personality annotations. We also propose a unified HHI representation named OpenHHI that is universally applicable to both generative and perceptive downstream tasks.}
   \label{fig:teaser}
\end{figure*}

\IEEEPARstart{T}{HE} capability to perceive and synthesize human-human interactions (HHI) is fundamental to the development of intelligent digital human systems, which have extensive applications across visual surveillance, AR/VR~\cite{ji2025towards}, interactive gaming, and robotics~\cite{liu2025takes,huang2026learning}. Nevertheless, modeling human-human dynamics remains profoundly challenging due to the inherent complexity and vast diversity of interaction patterns, coupled with severe mutual and self-occlusions. Although impressive strides have been made in both perception paradigms, \ieno, skeleton-based interaction recognition~\cite{pang2022igformer,duan2023skeletr}, and generative tasks, \ieno, action/text-conditioned interaction synthesis~\cite{xu2023actformer,interhuman}, existing methods often yield sub-optimal performance. This bottleneck primarily stems from the critical absence of a comprehensive benchmark capable of holistically capturing the multimodal nature of real-world HHIs.

The advancement of human-human interaction analysis is accompanied by the construction of dedicated interaction datasets~\cite{van2011umpm,sbu_kinect,chi3d,hi4d,interhuman}, as listed in~\cref{tab:dataset}. However, existing HHI datasets remain insufficient for comprehensive HHI modeling, exhibiting critical limitations at both the interaction data and multimodal annotation levels~\cite{van2011umpm,sbu_kinect,chi3d,hi4d,interhuman,baltruvsaitis2018multimodal,xu2023multimodal}.
At the data level, high-fidelity kinematic precision, diverse interaction patterns, dexterous hand articulations, and physically constrained modalities are indispensable. Previous attempts~\cite{hi4d,interhuman,mclean2025embody} estimating human motion via multi-view RGB cameras often suffer from estimation errors and artifacts, leading to inaccurate interaction modeling. Furthermore, small-scale datasets~\cite{van2011umpm,sbu_kinect,chi3d} inevitably restrict both data diversity and the range of interaction categories. The dexterous hand gestures play important roles for HHIs such as ``shaking hands'', ``grabbing'', ``waving'', \etc. However, to the best of our knowledge, there is no large-scale dataset providing high-fidelity finger movements for general human-human interactions. Additionally, most prior works overlook physically constrained interactions, which hold direct significance for downstream applications such as human-robot interaction.

Building upon the interaction data foundation, comprehensive multimodal annotations are also crucial yet underexplored in facilitating diverse downstream tasks, which encompass rich textual descriptions, interaction categories, causal interaction order, alongside interpersonal relationships and personalities. Text-driven HHI generation~\cite{interhuman,inter-x} has attracted much attention due to its significant practical applicability. Within this context, interaction categories can serve as the simplest semantic annotations, complemented by hierarchical textual descriptions that describe the interaction dynamics across multiple granularities. For instance, a single interaction sequence can be broadly described as a simple ``one person approaches the other and embraces her/him'', but it can also be richly detailed by incorporating part-level semantics. While simple descriptions offer a higher degree of semantic abstraction, fine-grained text annotations enable controllable interaction generation and better spatiotemporal alignment between motion and text modalities.
Recent works on human reaction generation~\cite{xu2024regennet} have demonstrated that interaction order annotations are of practical significance for delineating causal HHIs.
Moreover, the intimacy level and social relationships between individuals, together with their personalities, intuitively affect the interaction patterns, which should be considered for emotionally intelligent HHIs.

Beyond the aforementioned data and annotation limitations, fragmented interaction representations and inconsistent evaluation protocols are also significantly impeding the broader advancement of HHI analysis. Different works adopt different motion parameterizations and metric spaces, making experimental comparisons less reliable and often obscuring the actual impact of the representation itself. Moreover, the lack of a unified benchmark and standardized evaluation setting prevents a fair assessment of how different HHI models perform across generation and perception tasks.

To systematically address the identified limitations spanning data, annotations, representations, and evaluation protocols, we introduce Inter-X++ as a comprehensive benchmark to empower and advance versatile human-human interaction analysis, as illustrated in~\cref{fig:teaser}. 
Data acquisition leverages a novel hybrid motion capture system, combining an optical scheme for precise whole-body kinematics with an occlusion-robust inertial solution for dexterous hand gestures. The resulting dataset encompasses 40 daily interaction categories, featuring 11,388 sequences and over 8.1M frames from 89 distinct subjects.
On top of the data foundation, we further enrich the multimodal annotations with hierarchical fine-grained text descriptions, semantic interaction categories, causal interaction order, relationship and personality labels, as well as vertex-level contact and physically regularized interaction annotations, substantially broadening the utility of the benchmark.

With our proposed high-precision HHI data and the versatile annotations, we propose a unified testing ground comprising four categories of downstream tasks spanning generative and perceptive applications:
1) \textbf{Hierarchical texts} enable controllable text-driven human interaction generation~\cite{interhuman,inter-x} and interaction captioning tasks~\cite{guo2022tm2t,jiang2023motiongpt}, while simultaneously supporting a systematic investigation into the correlation between textual description granularity and the fidelity of generated motions;
2) \textbf{Interaction categories} facilitate action-conditioned human interaction generation~\cite{xu2023actformer} together with the human interaction recognition tasks~\cite{pang2022igformer,duan2023skeletr};
3) \textbf{Interaction order} enables the causal human reaction generation~\cite{chopin2023interaction,role_aware} and the causal interaction order inference tasks, \ieno, detecting the perpetrator in surveillance scenarios;
4) \textbf{Relationship and personality} make the stylized interaction generation~\cite{aberman2020unpaired} and the personality assessment possible.

To ensure rigorous benchmarking, we first systematically evaluate the impact of different interaction representations on experimental outcomes. Based on this analysis, we standardize our metric space by unifying all experiments using the continuous 6D rotation representation, while concurrently aligning the evaluation protocols across all eight downstream tasks. Furthermore, we present a single, unified HHI representation and model, termed OpenHHI, for both interaction understanding and generation tasks. Built upon the shared interaction representation, OpenHHI bridges multiple downstream tasks within a single cohesive modeling framework. Extensive experiments demonstrate the strong effectiveness of the proposed representation and model, yielding substantial gains across diverse tasks and establishing state-of-the-art performance on most evaluation metrics of the Inter-X++ benchmark.

Our contributions can be summarized as follows:
\begin{itemize}
  \item We collect a large-scale and most comprehensive human-human interaction dataset, featuring high-fidelity human body movements, diverse interaction patterns, expressive finger articulations and all rigorously augmented by vertex-level contact annotations and physical constraints.
  \item We complement Inter-X++ with hierarchical textual descriptions, semantic interaction categories, causal interaction order, relationship and personality, thereby empowering a broad spectrum of downstream applications;
  \item Building upon the multimodal annotations, we define four categories of distinct downstream tasks, with each explicitly featuring both generative and perceptive variants.
  \item We present a comprehensive analysis and unification of human-human interaction representations, alongside a standardized evaluation protocol for downstream tasks.
  \item We propose a unified human-human interaction representation called OpenHHI, which simultaneously facilitates both interaction understanding and generation tasks. Furthermore, it achieves substantial performance improvements across the proposed downstream tasks, establishing state-of-the-art results on multiple benchmarks.

\end{itemize}

A preliminary version of this work~\cite{inter-x} has been published in CVPR 2024 as a conference paper. In this extended version, we introduce substantial improvements in the following key aspects: \textbf{Firstly}, we drastically expand the annotation types and granularity of the existing dataset. We extend the original monolithic fine-grained textual annotations into hierarchical descriptions, enabling the investigation of how varying text granularities influence and control interaction motion generation. Furthermore, we upgrade the causal reasoning annotations to not only capture the interaction order of HHIs but also provide individual-specific textual descriptions, thereby facilitating a decoupled analysis of the two interacting subjects. Two novel modalities of vertex-level contact annotations and physically regularized HHI data are also supplemented to enhance the close-proximity and physical realism of the dataset and to support the development of physically aware interaction modeling.
\textbf{Secondly}, to circumvent the lack of standardized benchmarks arising from fragmented representations and evaluation protocols across the literature, we systematically analyze the impact of different interaction representations on experimental outcomes and unify the evaluation protocols. 
\textbf{Finally}, we propose OpenHHI, a unified HHI representation capable of simultaneously supporting both generative and perceptive tasks. The VQ-VAE-quantized interaction features are processed through a ViT encoder and subsequently channeled into a dual-branch decoder. One branch is dedicated to interaction reconstruction, while the other introduces a novel interaction captioning objective for semantic understanding. The joint optimization of these parallel branches effectively enforces the learning of a cohesive representation, which delivers remarkable performance gains across diverse downstream applications and sets multiple new state-of-the-art results.
Collectively, these systematic upgrades in HHI data, annotations, representations, and evaluation protocols establish a robust foundation for the continuous advancement of human-human interaction understanding and generation.

\begin{table*}[t]
  \centering
  \caption{\textbf{Dataset Comparisons.} A comprehensive comparison between Inter-X++ and existing human-human interaction datasets. \textbf{Sequences}: Total number of interaction clips; \textbf{Frames}: Total number of 3D motion frames; \textbf{Texts}: The number of textual annotations; \textbf{Scheme}: Motion data acquisition strategy; \textbf{Modality}: Kinematic representation (``Skel.'' denotes skeleton). Additionally, \textbf{Hands}, \textbf{Asyn.}, \textbf{Rel.\&Pst.}, \textbf{Physical}, and \textbf{Contact} indicate the inclusion of articulated hand gestures, asymmetry annotations, interpersonal relationships and personalities, physical plausibility, and explicit contact annotations, respectively. $^*$ denotes the integration of upgraded hierarchical language descriptions.}
  \label{tab:dataset}
  \begin{tabular}{lccccccccccc}
    \toprule
    Dataset & Year & Sequences & Frames & Texts & Scheme & Modality & Hands & Asyn. & Rel.\& Pst. & Physical & Contact\\
    \midrule
    UMPM~\cite{van2011umpm} & 2011 & 36 & 400K & \xmark & MoCap & Skel. & \xmark & \xmark & \xmark & \xmark & \xmark\\
    SBU Kinect~\cite{sbu_kinect} & 2012 & 300 & 7.5K & \xmark & RGB+D & Skel. & \xmark & \xmark & \xmark & \xmark & \xmark\\
    You2Me~\cite{ng2020you2me} & 2020 & 42 & 77K & \xmark & RGB+D & Skel. & \xmark & \xmark & \xmark & \xmark & \xmark\\
    NTU120~\cite{nturgbd120} & 2019 & 8,276 & 462K & \xmark & RGB+D & Skel. & \xmark & \xmark & \xmark & \xmark & \xmark\\
    Chi3D~\cite{chi3d} & 2020 & 373 & 63K & \xmark & MoCap & SMPL-X & \cmark & \xmark & \xmark & \xmark & \xmark\\
    ExPI~\cite{expi} & 2022 & 115 & 30K & \xmark & mRGB & Skel. & \xmark & \xmark & \xmark & \xmark & \xmark\\
    Hi4D~\cite{hi4d} & 2023 & 100 & 11K & \xmark & mRGB & SMPL & \xmark & \xmark & \xmark & \xmark & \cmark\\
    InterHuman~\cite{interhuman} & 2023 & 6,022 & 1.7M & 16,756 & mRGB & SMPL & \xmark & \xmark & \xmark & \xmark & \xmark\\
    ReMoCap~\cite{ghosh2023remos} & 2024 & 87 & 275.7K & \xmark & mRGB & Skel. & \cmark & \xmark & \xmark & \xmark & \xmark\\
    Inter-X~\cite{inter-x} & 2024 & 11,388 & 8.1M & 34,164 & MoCap & SMPL-X & \cmark & \cmark & \cmark & \xmark & \xmark\\
    InterDance~\cite{li2024interdance} & 2025 & - & - & \xmark & MoCap & SMPL-X & \cmark & \xmark & \xmark & \xmark & \xmark \\
    Embody 3D~\cite{mclean2025embody} & 2025 & - & - & \cmark & mRGB & SMPL-X & \cmark & \xmark & \xmark & \xmark & \xmark\\
    \midrule
    Inter-X++ & 2026 & \textbf{11,388} & \textbf{8.1M} & \textbf{102,492$^*$} & MoCap & SMPL-X & \cmark & \cmark & \cmark & \cmark & \cmark\\
    \bottomrule
  \end{tabular}
\end{table*}

\section{Related work}

In this section, we organize the related works into three aspects. We begin with an overview of existing human motion and human-human interaction (HHI) datasets in~\cref{sec:related_1}. We then systematically review the major HHI tasks with their developments in~\cref{sec:related_2}. Finally, we summarize the evaluation protocols and benchmarking settings adopted in HHI research in~\cref{sec:related_3}.

\subsection{Human Motion and Interaction Datasets}
\label{sec:related_1}

Compared to RGB video streams, 3D human motion representations provide a high-level, efficient, privacy-friendly, and illumination-invariant modality~\cite{xu2022skeleton}. The evolution of human motion datasets~\cite{ionescu2013human3,nturgbd120,mahmood2019amass,motion-x,zhang2025motion,motion_survey} has fundamentally driven progress in human motion understanding, with early large-scale benchmarks primarily relying on discrete action labels~\cite{nturgbd120,punnakkal2021babel}, laying the foundation for discriminative motion modeling. To capture fine-grained semantics and facilitate open-vocabulary analysis, subsequent datasets incorporated rich natural language descriptions~\cite{humanml3d,motion-x,zhang2025motion}, enabling cross-modal alignment between kinematics and text. Beyond isolated motions, recent efforts emphasize context-aware modeling to address real-world complexities. Specifically, datasets featuring synchronized audio signals~\cite{li2021learn} have been introduced for rhythmic and speech-driven behavior analysis. Simultaneously, the proliferation of datasets with explicit 3D scene and object geometries~\cite{taheri2020grab,hassan2021stochastic,wang2022humanise} provides crucial physical grounding, which is indispensable for advancing robust human-environment interactions and real-world human-centric tasks.

Beyond single-person motion capture, a variety of human-human interaction datasets have been introduced~\cite{sbu_kinect,chi3d,hi4d,interhuman,ghosh2023remos,li2024interdance,mclean2025embody}, encompassing diverse scales, modalities, and functionalities, as shown in~\cref{tab:dataset}. Notably, the recent InterHuman dataset~\cite{interhuman} significantly advances this domain by providing large-scale HHI sequences paired with semantic text. While existing benchmarks typically compromise on kinematic precision or holistic detail, our preliminary work introduced the Inter-X dataset~\cite{inter-x} to address these limitations through high-fidelity motions, fine-grained text, and synchronized hand articulation. In this extension, we significantly elevate Inter-X by incorporating a richer spectrum of modalities. Specifically, we augment the dataset with hierarchical textual descriptions, detailed contact annotations, and enhanced physical realism, thereby establishing a substantially more comprehensive and physically grounded foundation for HHI modeling.

\subsection{Human-Human Interaction Tasks}
\label{sec:related_2}

Human-human interaction (HHI) constitutes a fundamental component of social behavior understanding and plays a critical role in a wide range of human-centric applications, including virtual and augmented reality~\cite{ji2025towards} and human-robot interaction~\cite{liu2025takes,chen2025rhino,huang2026learning}.

\subsubsection{HHI Generation}
Human motion generation aims to synthesize kinematic sequences that are both physically plausible and behaviorally diverse, conditioned on various control modalities. Recent advancements have predominantly focused on single-person generation driven by discrete action labels~\cite{action2motion,actor,mdm,xu2023actformer}, free-form textual descriptions~\cite{petrovich22temos,humanml3d,motion-x,lu2023humantomato,motiondiffuse}, and acoustic signals~\cite{Ao2023GestureDiffuCLIP}. Beyond the synthesis of isolated individuals, recent efforts have further expanded toward complex multi-person interactions~\cite{xu2023actformer,interhuman,wu2025intermamba,geng2026disentangled}. in2In~\cite{ruiz2024in2in} and ComMDM~\cite{commdm} leverage pre-trained motion priors derived from single-person motion generation to facilitate two-person interaction synthesis. TIMotion~\cite{wang2025timotion} jointly accounts for temporal modeling and interaction mixing to achieve efficient HHI generation. InterMask~\cite{javed2024intermask} employs a VQ-VAE to discretize individual motions into 2D tokens, modeling HHIs through a generative masked modeling framework. Human-X~\cite{ji2025towards} systematically addresses real-time responsiveness, physical feasibility, and safety constraints in human interactions to empower immersive human-machine collaboration. Anchored on proximal interactive poses for versatile interaction animation, Ponimator~\cite{liu2025ponimator} successfully transfers interaction knowledge from motion capture data to open-world scenarios. Furthermore, Interact2Ar~\cite{ruiz2025interact2ar} utilizes an autoregressive diffusion model to synthesize full-body HHIs, naturally encompassing fine-grained hand articulations.

\begin{figure*}[t]
  \centering
   \includegraphics[width=1.0\linewidth]{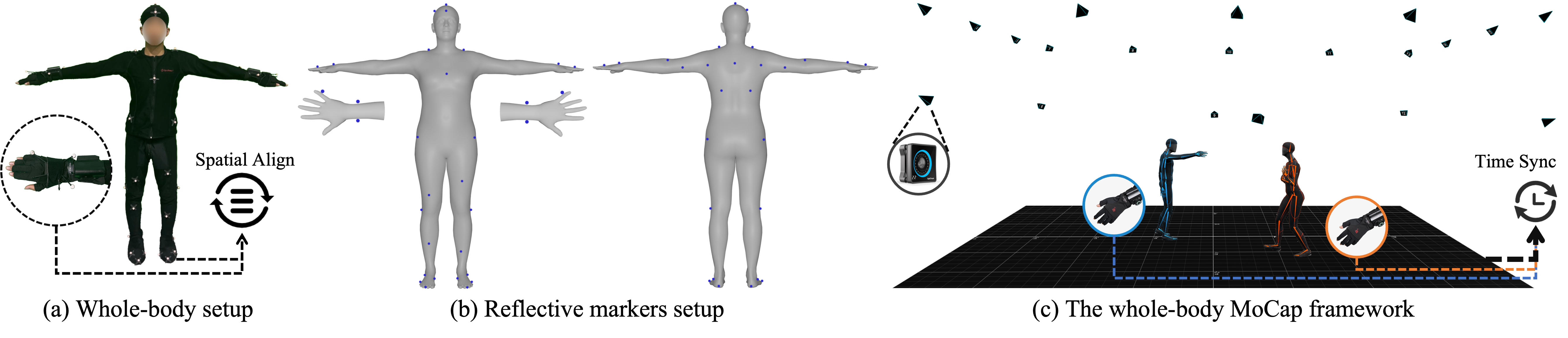}
   \caption{\textbf{The Inter-X++ whole-body hybrid MoCap system.} (a). Spatial integration and alignment of the optical MoCap suit and the IMU-based gloves is achieved via a rigid triangular bracket of reflective markers; (b). Detailed view of the reflective marker configuration; (c). Temporal synchronization of the human body and hand kinematics within the unified whole-body motion capture system.}
   \label{fig:setup}
\end{figure*}

\subsubsection{Human Reaction Generation}

Emerging works have begun to explore human reaction generation, aiming to synthesize plausible responsive behaviors conditioned on antecedent human motions~\cite{ghosh2023remos}, or more generalized inputs such as video sequences~\cite{yu2025hero,zhang2025egoreact} and interlocutor speech~\cite{luo2026reactmotion}. Specifically, InterFormer~\cite{chopin2023interaction} introduces an interaction transformer architecture equipped with both temporal and spatial attention mechanisms for human reaction generation. HIG~\cite{role_aware} proposes an annotation-free approach to automatically learn the underlying dynamics between actors and receivers. To ensure physical realism, PhysReaction~\cite{liu2024physreaction} focuses on generating physically plausible human reactions in real time, while ARFlow~\cite{jiang2025arflow} directly models the continuous action-to-reaction mapping via flow matching~\cite{lipman2022flow}. Notably, ReGenNet~\cite{xu2024regennet} constructs the first multi-setting human-reaction generation framework, incorporating explicit annotations for interaction orders. Extending beyond pure human-human scenarios, iHuman~\cite{liu2023interactive} establishes an online reaction generation system for human-humanoid interactions. Addressing real-time interaction constraints, Ready-to-React~\cite{cen2025ready} formulates an online policy that generates reactive motions for each individual based on past observations, and ReMoGen~\cite{ye2026remogen} synthesizes real-time, high-quality, and coherent reactions from heterogeneous interaction data. Furthermore, HERO~\cite{yu2025hero} extracts reaction generation priors directly from videos, encompassing human-human, human-animal, and human-scene interactions. Similarly, EgoReAct~\cite{zhang2025egoreact} processes streaming egocentric videos to synthesize spatially grounded and realistic human reactive motions. ReactMotion~\cite{luo2026reactmotion} introduces a novel task of generating listener body reactions driven by speaker utterances. By providing explicit temporal-order annotations and equipping each participant with participant-specific textual descriptions, Inter-X++ further advances the development of human reaction generation.

\subsubsection{HHI Recognition}
Skeleton-based human action recognition has been extensively investigated~\cite{stgcn,2s-agcn,ms-g3d,ctrgcn,hdgcn}. As a challenging sub-field, human interaction recognition~\cite{pang2022igformer,duan2023skeletr,wang2023human,wu2024sportshhi,wei2024nonverbal,li2026multiple} further demands precise modeling of semantic and spatiotemporal correlations among individuals. Beyond discrete action classification, human kinematics inherently encode rich biometric and psychological cues~\cite{vinciarelli2014survey}. For instance, gait recognition~\cite{wan2018survey} seeks to identify individuals through distinct motion patterns, while other studies~\cite{durupinar2016perform,delgado2022automatic} leverage body movements as predictive indicators of personality traits. Featuring large-scale action-motion and text-motion alignments, our Inter-X++ dataset directly drives advancements in human action recognition. More importantly, it establishes a novel foundation for quantitatively assessing interpersonal relationships and individual personalities from interactive motions.

\subsubsection{HHI Reconstruction}

Another research direction focuses on recovering high-precision HHIs from images or videos~\cite{fang2022alphapose}. Hi4D~\cite{hi4d} provides a robust benchmark tailored for close human interactions, featuring rich, interaction-centric annotations. CloseMoCap~\cite{shuai2023reconstructing} and AvatarPose~\cite{lu2024avatarpose} attempt to reconstruct close-contact interaction sequences from multiple calibrated cameras. Addressing the more challenging monocular setting, MultiPhys~\cite{ugrinovic2024multiphys} successfully recovers robust and physically aware multi-person interactive motions. To mitigate the severe occlusions inherent in interacting scenarios, Dual-Human~\cite{fang2024capturing} leverages semantic reaction priors to assist in inferring occluded regions, thereby significantly enhancing reconstruction quality. Furthermore, CloseApp~\cite{huang2025reconstructing} introduces a dual-branch optimization framework that exploits human appearance, proxemics, and physical constraints to effectively resolve visual ambiguities.

\subsection{HHI Representation and Benchmarking}
\label{sec:related_3}
Human motion and HHI analysis remain hindered by fragmented representations and the lack of a unified evaluation protocol. Early works such as HumanML3D~\cite{humanml3d} adopted carefully designed yet redundant kinematic representations, and HumanTOMATO~\cite{lu2023humantomato} further extended this formulation to whole-body motion. More recent frameworks, including STMC~\cite{petrovich2024multi} and MotionStreamer~\cite{xiao2025motionstreamer}, have introduced substantially different representational formats. A similar issue persists in HHI modeling: InterGen~\cite{interhuman} follows the HumanML3D formulation and defines both canonical and non-canonical representations, whereas Inter-X~\cite{inter-x} adopts a direct 6D rotation parameterization. This inconsistency undermines fair comparison and benchmark rigor. To address it, we analyze how motion representations affect generation performance and establish a unified protocol based on a standardized representation for fair and consistent benchmarking.

\section{The Inter-X++ Dataset}

We introduce Inter-X++, a large-scale dataset designed for comprehensive multimodal human-human interaction analysis. 
In this section, we first introduce the data capturing system with postprocessing to curate human-human interaction data in~\cref{sec:data_capture}. We then detail the dataset composition, including motion data, action category labels, hierarchical textual descriptions, interaction order, contact annotations, and relationship and personality attributes of the subjects in~\cref{sec:data_taxonomy}. These annotations significantly enrich the modality spectrum of human-human interaction data and support a more detailed understanding and modeling of human-human interactions.

\subsection{Dataset Capturing System}
\label{sec:data_capture}

For interaction data acquisition, we impose two key requirements on the capturing setup: ensuring high-fidelity motion capture and enabling the capture of dexterous finger movements, which are crucial for a wide range of daily interaction scenarios such as handshaking and rock-paper-scissors.
Many existing human motion datasets rely on multi-view RGB-based technologies~\cite{nturgbd120,interhuman} to extract human motions from RGB videos. Although such pipelines preserve the natural image information, these datasets suffer from severe occlusions and penetrations (especially for close human-human interactions), and the subtle finger movements are also difficult to obtain precisely. 
Other approaches employ inertial MoCap suits to record human motions~\cite{lu2025humoto}. Despite their convenience and lack of environmental constraints, our experiments show that error accumulation can introduce substantial inaccuracies in multi-person interaction capture, particularly in the relative displacement between individuals.
Thus, we choose the optical motion capture system for its high precision to obtain the body movements.
Additionally, we further employ inertial gloves to capture the finger gestures, which are inherently robust to hand flipping, rapid hand motions and occlusions.
The overview of the dataset capturing system is illustrated in~\cref{fig:setup}.

We deploy the OptiTrack MoCap system~\cite{optitrack} with 20 PrimeX 22 infrared cameras recording at the resolution of 2048$\times$1088 and 120 FPS. The dimensions of our MoCap venue are 8.5 meters in length, 5.4 meters in width, and 3.3 meters in height, which are sufficient to cover most daily human-human interactions. The optical motion capture scheme ensures an error of $\pm0.15$ mm, which is significantly lower than that of the multi-view RGB cameras or inertial MoCap suits. More importantly, the system demonstrates strong robustness in capturing multi-person interactions, especially for poses involving close physical contact, and remains reliable even when some reflective markers are occluded.
To capture the dexterous hand gestures and achieve robustness to hand rotations, rapid movements, and occlusions, we adopt the inertial solution of the commercial Noitom Perception Neuron Studio (PNS) gloves~\cite{noitom}. The subtle finger movements can be captured in real-time, disregarding the self-occlusion and occlusion with the other person during the interactions.

For each capture session, two volunteers are randomly paired to wear the OptiTrack MoCap suits with 41 reflective markers and the inertial gloves as depicted in~\cref{fig:setup}(a),(b). Both of them are carefully calibrated for the hybrid optical-inertial setup, respectively, before they perform the interactions. We provide timecodes for the OptiTrack MoCap system and the PNS gloves using Tentacle Sync so that the body and hands can be temporally synchronized. 
For each recording batch, we arrange five action categories with five repetitions to ensure variability, which improves efficiency and ensures the continuity of the volunteers' actions. Taking handshaking as an example, the two participants are encouraged to perform different handshaking patterns, such as single-hand and double-hand handshakes, varying the interpersonal distance, or interacting while standing or sitting on chairs, in order to increase the diversity of the dataset. The volunteers pause for several seconds between two interaction snippets to improve capture quality and ease the subsequent temporal segmentation.
We also recalibrate the PNS inertial gloves frequently between recording batches to mitigate the error accumulation.

\begin{figure}[t]
  \centering
   \includegraphics[width=1.0\linewidth]{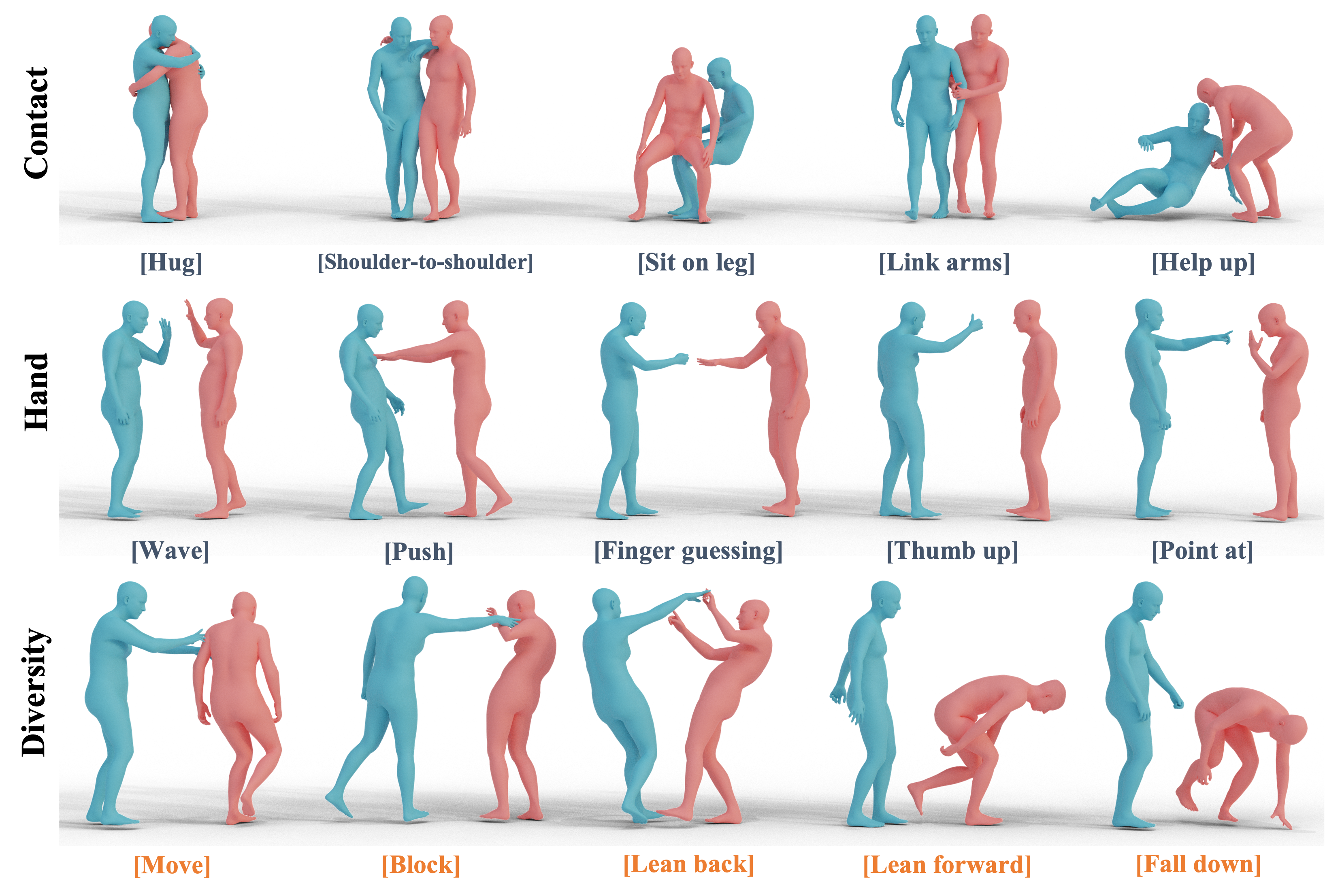}
   \caption{\textbf{Motion characteristics of Inter-X++.} Our proposed Inter-X++ dataset is distinguished by the 1) Precise contact for close HHIs; 2) Dexterous hand gestures; 3) Diverse action and reaction patterns.}
   \label{fig:dataset_characteristics}
\end{figure}

\subsection{Dataset Taxonomy}
\label{sec:data_taxonomy}

This subsection first presents the processing pipeline for the captured human-human interaction data and then introduces the procedures for further enriching the data with multimodal annotations, yielding 11,388 pairs of SMPL-X~\cite{smplx} interaction sequences, 40 semantic action categories with diverse action/reaction patterns, 102,492 hierarchical text descriptions, interaction order labels, contact annotations and the relationship for 59 groups and personality for 89 volunteers. Representative characteristics of the Inter-X++ dataset are illustrated in~\cref{fig:dataset_characteristics}.

\subsubsection{Interaction Data}
To obtain the full-body human motion data, the crux of the postprocessing is the alignment between the body poses from the OptiTrack MoCap system and the finger gestures from the inertial gloves.
Temporally, we retrieve the overlapping intervals between the body pose and the hand pose sequences based on the millisecond-level timestamps. Spatially, these two modalities are naturally integrated through the shared wrist rotation obtained from the triangular locating bracket attached to the hand dorsum.
Following the temporal and spatial alignment, we derive the whole-body HHI skeleton sequences. Since the dataset is collected in batches, we further perform temporal segmentation to ensure that each sequence corresponds to an atomic interaction. We recruit annotators to manually identify the start and end frames of each atomic interaction. The segmentation results are carefully collected and verified, and the long batch-captured sequences are then trimmed into atomic segments accordingly.

We adopt the SMPL-X~\cite{smplx} parametric model due to its high expressiveness in capturing intricate human body poses and fine-grained hand articulations, as well as its broad applicability to downstream tasks. Formally, a sequence of $N$ frames is parameterized by body pose parameters $\theta\in\mathbb{R}^{N\times55\times3}$, shape parameters $\beta\in\mathbb{R}^{N\times10}$ and translation parameters $t\in\mathbb{R}^{N\times3}$. The shape parameters $\beta$ are initialized using the height and weight of each volunteer, following the methodology in~\cite{virtual_caliper}. Subsequently, a carefully tuned optimization algorithm is employed to fit the SMPL-X parameters to the captured MoCap keypoints by minimizing the following objective function:
\begin{equation}
  E(\theta,t)=\lambda_1\frac{1}{N}\sum_{j \in \mathcal{J}}\lambda_{p}||\bm{J}_j(\mathbb{M}(\theta,t))-\bm{g}_j||_2^2+\lambda_2||\theta||_2^2,
\end{equation}
where $\mathcal{J}$ denotes the joint set, $\mathbb{M}$ represents the SMPL-X parametric model, $\bm{J}_j$ is the regressor function for joint $j$, $\bm{g}$ indicates the skeleton captured by the MoCap system. The terms $\lambda_1$, $\lambda_2$ and $\lambda_p$ are different optimization weights, with $\lambda_p$ specifically designed to assign varying degrees of importance to different body parts. Please refer to the supplementary materials for comprehensive implementation details.

\subsubsection{Interaction Categories}
We curate the interaction categories by drawing upon the established HHI datasets~\cite{chi3d,nturgbd120,interhuman} and leveraging large language models~\cite{gpt3}. Ultimately, we figure out 40 daily interaction categories, representing, to the best of our knowledge, one of the most comprehensive behavioral taxonomies in the HHI domain.
To ensure data quality, volunteers are instructed to execute the interaction both \textit{naturally} and \textit{diversely}.
Data collection is conducted in a controlled, enclosed MoCap studio environment. Volunteers are instructed to remain relaxed and perform the interactions according to their habitual behavioral patterns. To ensure high-fidelity data capture and prevent fatigue, a brief rest period of several seconds was allocated between consecutive trials.
Meanwhile, the diversity is manifested across three dimensions: 1) action variations, \eg, raising the left, right, or both hands for ``raising hands''; 2) reaction variations, \eg, resisting, stepping back, or falling when ``being pushed''; and 3) heterogeneous initial body states, \eg, standing, sitting, crouching, or lying down.
Furthermore, each interaction is performed five times to maximize intra-class variability.

\subsubsection{Hierarchical Text Descriptions}

Natural language exhibits inherent flexibility, allowing for varying levels of abstraction and granularity. For the same interaction sequence, the descriptions may vary as ``Handshake with each other'' or ``Shake hands with their \textit{right hands}'', where the latter specifies particular body parts. Highly complex and fine-grained textual descriptions yield more precise motion representations and facilitate finer control over interaction synthesis; conversely, simpler text prompts offer a higher degree of semantic abstraction.
Inter-X~\cite{inter-x} originally provided highly detailed textual annotations with extensive human body part-level descriptions. Specifically, we develop an annotation tool based on~\cite{ait-viewer}, allowing annotators to freely manipulate the 3D viewpoint, including scaling and 360-degree rotation to observe interaction details from any perspective. For every interaction sequence, three independent annotators are tasked with providing fine-grained descriptions encompassing: 1) coarse full-body movements, 2) finger articulations, and 3) relative spatial orientations between the participants. The raw textual data was subsequently refined using GPT-3.5~\cite{gpt3} to rectify typographical errors and validated through manual spot-checking. Consequently, the resulting descriptions average $\sim$35 words, substantially exceeding prior benchmarks and underscoring the exceptional semantic density of our dataset.

However, we empirically find that the state-of-the-art interaction generation models still struggle to achieve consistent alignment between the textual inputs and the synthetic interactions. Wu~\etal~\cite{wu2025mg} points out that overly long, detailed descriptions may distract the generative model from capturing the global semantics of the motions. To achieve a higher degree of semantic abstraction, we uniformly render 12 frames per interaction sequence and provide these visual inputs, coupled with the three corresponding raw detailed descriptions, to GPT-5.1~\cite{gpt5.1} for text simplification, thereby synthesizing a concise version of the action narrative. We further define the action category as the most simplified textual representation. Ultimately, providing texts with hierarchical levels of abstraction empowers diverse conditional motion generation tasks and their corresponding evaluations.

\subsubsection{Causal Interaction Order}
Exploring the causal relationships inherent in social HHI behaviors, where the action of one individual precipitates a reaction from another, is crucial for advancing the understanding of HHIs~\cite{sbu_kinect}. To model this, volunteers are asked to explicitly annotate the causal order of the active and reactive participants within every atomic interaction clip.
For highly symmetric interactions, such as handshaking, the individual who extends their hand first is typically designated as the actor. 

Furthermore, beyond merely determining the causal order for each interaction clip, we aim to acquire more fine-grained and role-specific action descriptions for both the actor and the reactor, respectively. To achieve this, we uniformly sample 12 rendered frames from each interaction sequence alongside the initial global textual descriptions, and then feed them into GPT-5.1~\cite{gpt5.1}, prompting the model to generate distinct action descriptions for each participant. By providing distinct action narratives for each individual, we effectively augment the holistic HHI descriptions, enabling more fine-grained, text-driven control over individual participants.

\subsubsection{Contact Annotations}

The annotation of contact regions is crucial for a deeper understanding of close human-human interactions. However, acquiring precise contact areas remains highly challenging. Leveraging our high-precision OptiTrack optical MoCap system, we directly employ a simple yet efficient distance-based method to compute these regions. Specifically, we define the contact area as locations where the inter-subject distance is less than 0.05\,m, thereby providing fine-grained, vertex-level contact annotations.

\subsubsection{Relationship \& Personality}

Exploring the correspondence between HHI and their relationship \& personality is a niche~\cite{durupinar2016perform,delgado2022automatic} where the essence lies in the disentanglement of the personality factors from body motions. We employ the standard Big-Five Personality Model~\cite{wiggins1996five}, utilizing the NEO Five-Factor Inventory~\cite{mccrae2004contemplated} to evaluate their personalities of openness, conscientiousness, extraversion, agreeableness, and neuroticism. Additionally, participants assessed their mutual familiarity on a four-level scale and identified their social relationships into five types: strangers, friends, romantic partners, schoolmates, or family.

\subsubsection{Physical Plausibility}
To enhance the physical plausibility of Inter-X++, we input the kinematics-based motions into a physics simulator and utilize the imitation policy PHC~\cite{PHC} trained with reinforcement learning to drive humanoid agents to imitate these motions, thus correcting the motions to conform to physical constraints. However, directly using the original PHC trained only on AMASS results in insufficient success rates and motion jittering during imitation. Therefore, we fine-tuned PHC on Inter-X++. Similar to MultiPhys~\cite{ugrinovic2024multiphys}, the policy is trained in a single-agent environment during fine-tuning, which helps correct physical artifacts such as foot sliding. During the evaluation phase, we correct unreasonable interactions through having different agents sharing the same simulation environment simultaneously, allowing the physics engine to directly apply physical constraints between them, such as preventing interpenetration, thus solving physical artifacts in multi-person interactions. Finally, the corrected motions are converted back to the SMPL-X to achieve more accurate and physically plausible multi-person motions.

\section{Task Taxonomy}

The high-fidelity HHI sequences within our dataset, characterized by complex hand articulations, introduce unprecedented challenges to current benchmarks. Leveraging our comprehensive and versatile annotations, we further formulate several novel downstream tasks oriented toward real-world applications. Formally, a given HHI sequence is denoted as $\bm{m}=\langle\bm{x}, \bm{y}\rangle$, paired with its corresponding annotations: the discrete action category $\bm{l}_a$, textual description $\bm{l}_t$, causal interaction order $\bm{l}_c$, interpersonal relationship $\bm{l}_r$, and individual personalities $\bm{l}_p=\langle\bm{l}_{p_x}, \bm{l}_{p_y}\rangle$.

\subsection{Text-Related Tasks}

\paragraph{Text-conditioned HHI generation} 
While text-driven single-person motion generation has been extensively explored across various benchmarks~\cite{humanml3d,motion-x}, extending this capability to multi-person scenarios remains challenging. Equipped with hierarchical textual annotations, our dataset unlocks novel paradigms for controllable HHI generation. Concurrently, it introduces unprecedented challenges, particularly in synthesizing dexterous hand articulations and precisely aligning part-level textual descriptions with interactive kinematics. We also systematically evaluate the generation performance by conducting comparative experiments driven by hierarchical textual descriptions of varying granularity. Formally, this task is defined as learning a generative mapping function $F_{t2m}$:
\begin{equation}
  F_{t2m}(\bm{l}_t) \mapsto \bm{m}.
  \label{eq:t2m}
\end{equation}

\paragraph{HHI captioning}
Beyond conventional discrete action recognition, HHI captioning has emerged as a novel task~\cite{guo2022tm2t,jiang2023motiongpt} that aims to synthesize comprehensive natural language descriptions directly from HHI sequences. This paradigm fundamentally reinforces the cross-modal alignment between complex kinematics and text, enabling the automated generation of diverse and semantically coherent captions. Formally, this task can be formulated as:
\begin{equation}
  F_{m2t}(\bm{m}) \mapsto \bm{l}_t.
  \label{eq:m2t}
\end{equation}

\subsection{Action-Related Tasks}

\paragraph{Action-conditioned HHI generation}
Conditioned on a discrete action label, the generative mapping $F_{a2m}(\cdot)$ aims to synthesize diverse and physically plausible HHI sequences~\cite{xu2023actformer,actor,mdm}. Empowered by our Inter-X++ dataset, this task is substantially elevated to model high-fidelity interactive behaviors that encompass dexterous finger kinematics. Formally, it is defined as:
\begin{equation}
  F_{a2m}(\bm{l}_a) \mapsto \bm{m}.
  \label{eq:a2m}
\end{equation}

\paragraph{HHI recognition}
Skeleton-based HHI recognition is essential for downstream applications such as visual surveillance~\cite{pang2022igformer,duan2023skeletr}. We posit that the integration of fine-grained hand articulations will significantly enhance the discriminative capacity of existing recognition frameworks. Formally, this task is defined as:
\begin{equation}
  F_{m2a}(\bm{m}) \mapsto \bm{l}_a.
  \label{eq:m2a}
\end{equation}

\subsection{Interaction-Order-Related Tasks}

\paragraph{Human reaction generation}
Despite its extensive applicability in immersive AR/VR and interactive gaming, human reaction synthesis~\cite{chopin2023interaction,role_aware} remains a comparatively underexplored domain. The explicit annotation of actor-reactor causal orders provides a critical foundation for investigating role asymmetry and precisely modeling responsive dynamics in HHIs. Formally, this process is defined as:
\begin{equation}
  F_{c2m}(\bm{l}_c, \bm{l}_a, \bm{x}) \mapsto \bm{y}.
  \label{eq:c2m}
\end{equation}

\paragraph{Causal interaction order inference}
The causal interaction order inference mapping, denoted as $F_{m2c}(\cdot)$, aims to explicitly distinguish the active instigator (actor) from the passive responder (reactor) within a given interaction sequence. This capability significantly benefits downstream applications such as intelligent surveillance and automated sports analytics:
\begin{equation}
  F_{m2c}(\bm{m}) \mapsto \bm{l}_c.
  \label{eq:m2c}
\end{equation}

\subsection{Relationship- and Personality-Related Tasks}

\paragraph{Stylized HHI generation}
Interpersonal relationships and individual personality traits offer powerful, high-level stylization priors for customized interaction synthesis. The substantial diversity of participants in our dataset, coupled with extended motion sequences for each subject, robustly supports the learning of these stylistic nuances. Formally, this generation task is formulated as:
\begin{equation}
  F_{s2m}(\bm{l}_r, \bm{l}_p, \bm{l}_a) \mapsto \bm{m}.
  \label{eq:s2m}
\end{equation}

\paragraph{HHI-based personality assessment}
While prior studies~\cite{durupinar2016perform,delgado2022automatic} have established body movements as viable predictors of individual personality, we leverage our dataset to propose the novel task of joint personality and relationship assessment. This quantitative evaluation of behavioral kinematics holds substantial potential for disciplines such as education, medical diagnostics, and sports analytics. Specifically, the assessment mapping is defined as:
\begin{equation}
  F_{m2s}(\bm{m}) \mapsto \{\bm{l}_r, \bm{l}_p\}.
  \label{eq:m2s}
\end{equation}

\section{OpenHHI: Unified Representation and Model}

To establish a more effective interaction representation and a rigorous evaluation benchmark, we first conduct a systematic study of widely used interaction representations in~\cref{sec:hhi_rep}. Then, we introduce a unified interaction encoder which can serve both perceptive and generative tasks in~\cref{sec:unified_rep}.

\subsection{Human-Human Interaction Representation}
\label{sec:hhi_rep}

The heterogeneity of interaction representations across existing HHI methods confounds cross-method comparisons, making it difficult to disentangle improvements attributable to model design from those arising from the representation itself. To support systematic studies and standardized evaluation of HHI models, we provide the same interaction motions in five representative forms: axis-angle, quaternion, rotation matrix, continuous 6D rotation~\cite{rot_6d}, and the non-canonical representation adopted by InterGen~\cite{interhuman}. We will analyze these representations to identify the best-performing choice, which is then used consistently throughout all subsequent experiments to ensure fair and uniform comparisons.

\subsubsection{Rotation representations}
We adopt the original SMPL-X motions and represent the joint rotations using axis-angle, quaternion, rotation matrix, or continuous 6D rotation~\cite{rot_6d}. In addition to the rotation features, each representation includes the global root translation $\bm{t}$ and its acceleration $\dot{\bm{t}}$. Taking the 6D rotation representation as an example, the motion feature is formulated as
\begin{equation}
  \bm{x}^{\mathrm{6D}}
  =
  [\bm{r}^{\mathrm{6D}},\bm{t},\dot{\bm{t}}],
  \label{eq:rot_rep}
\end{equation}
where $\bm{r}^{\mathrm{6D}}$ contains the 6D rotations of all SMPL-X joints. The four representations therefore share the same global trajectory information and differ only in their rotation parameterization.

\subsubsection{Non-canonical representation}
Following InterGen~\cite{interhuman}, we consider the non-canonical representation for HHI as:
\begin{equation}
  \bm{x}
  =
  [\bm{j}_{g}^{p},
   \bm{j}_{g}^{v},
   \bm{j}^{r},
   \bm{c}^{f}],
  \label{eq:noncanonical_rep}
\end{equation}
where $\bm{j}_{g}^{p}$ and $\bm{j}_{g}^{v}$ denote global joint positions and velocities, $\bm{j}^{r}$ denotes local joint rotations in the 6D rotational format, and $\bm{c}^{f}$ denotes binary foot-ground contact states.

\subsection{Unified Interaction Encoder}
\label{sec:unified_rep}

Generative and perceptive HHI tasks place complementary demands on interaction features. Generation requires a representation that retains fine-grained body poses, temporal dynamics, and interpersonal contact geometries so that the original interaction can be faithfully recovered, whereas understanding favors compact features organized by high-level interaction semantics. Training separate encoders for the two regimes duplicates computation and prevents the two forms of supervision from benefiting each other. We therefore introduce a unified HHI representation that learns a single representation for both HHI generation and understanding. Its core design consists of an interaction VQ-VAE, a shared ViT encoder, and two complementary training branches for interaction reconstruction and semantic understanding.

\subsubsection{Unified interaction tokenizer}
Given a paired interaction $\bm{m}$ in the 6D rotational representation, we first use a VQ-VAE to compress the high-dimensional motion sequence into a compact motion-aware latent space:
\begin{equation}
  \bm{z}_{q}
  =
  \mathcal{Q}\!\left(\mathcal{E}_{\mathrm{vq}}(\bm{m})\right),
  \qquad
  \bm{h}
  =
  \mathcal{E}_{\mathrm{vit}}(\bm{z}_{q}),
  \label{eq:openhhi_encoder}
\end{equation}
where $\mathcal{E}_{\mathrm{vq}}$ and $\mathcal{Q}$ denote the VQ-VAE encoder and codebook quantizer, respectively, and $\mathcal{E}_{\mathrm{vit}}$ is the shared interaction encoder. The output $\bm{h}$ serves as the unified HHI representation for both perceptive and generative tasks. During the training phase, we employ two branches for simultaneous supervision, comprising a reconstruction branch and an understanding branch. These two branches are entirely separate, and we will introduce them in detail below.

\subsubsection{Interaction reconstruction branch}
The reconstruction branch encourages $\bm{h}$ to retain the detailed kinematics required by generation. A ViT decoder $\mathcal{D}_{\mathrm{rec}}$ maps the unified representation back to the quantized latent space, and the VQ-VAE decoder $\mathcal{D}_{\mathrm{vq}}$ reconstructs the paired interaction:
\begin{equation}
  \begin{aligned}
  \widehat{\bm{z}}_{q}
  &=
  \mathcal{D}_{\mathrm{rec}}(\bm{h}),
  &\widehat{\bm{m}}
  &=
  \mathcal{D}_{\mathrm{vq}}(\widehat{\bm{z}}_{q}).
  \end{aligned}
  \label{eq:openhhi_reconstruction}
\end{equation}
Following the standard VQ-VAE reconstruction objective, we supervise this branch solely in the motion space:
\begin{equation}
  \mathcal{L}_{\mathrm{rec}}
  =
  \ell_{1}(\bm{m},\widehat{\bm{m}}),
  \label{eq:openhhi_rec_loss}
\end{equation}
where $\ell_{1}$ denotes the element-wise reconstruction loss between the input and reconstructed interaction motions.

\subsubsection{Interaction understanding branch}
Reconstruction alone mainly focuses on low-level kinematic fidelity and provides limited supervision for learning the semantics of an interaction. We therefore introduce an interaction captioning branch that uses the fine-grained textual descriptions as semantic supervision. Specifically, an autoregressive text decoder takes the unified interaction representation $\bm{h}$ as its condition and predicts the corresponding description token by token. The understanding objective is defined as the captioning loss:
\begin{equation}
  \mathcal{L}_{\mathrm{und}}
  =
  -\sum_{k=1}^{K}
  \log p\!\left(
  w_{k}\mid w_{<k},\bm{h}
  \right),
  \label{eq:openhhi_und_loss}
\end{equation}
where $w_k$ is the $k$-th caption token and $w_{<k}$ denotes the preceding tokens. By requiring $\bm{h}$ to support interaction description generation, this branch encourages the unified representation to capture high-level action semantics in addition to motion details.

\begin{table*}[t]
  \centering
  \caption{Experimental comparison of the human-human interaction representations on Inter-X++. We evaluate five interaction representations under a unified experimental protocol. $\pm$ indicates 95\% confidence interval and $\rightarrow$ means the closer the better. \textbf{Bold} indicates best results and \underline{underline} indicates second best results.}
  \label{tab:ablation_data_rep}
  \resizebox{1\textwidth}{!}{
  \begin{tabular}{lccccccc}
    \toprule
    \multirow{2}{*}{Method}  & \multicolumn{3}{c}{R Precision$\uparrow$} & \multirow{2}{*}{FID $\downarrow$} & \multirow{2}{*}{MM Dist$\downarrow$}  & \multirow{2}{*}{Diversity$\rightarrow $} & \multirow{2}{*}{MModality $\uparrow$}\\
    \cmidrule(lr){2-4} & Top 1 & Top 2  & Top 3 \\
    \midrule
      Real & $0.448^{\pm0.003}$ & $0.639^{\pm0.003}$ & $0.743^{\pm0.003}$ & $0.001^{\pm0.0001}$ & $3.652^{\pm0.009}$ & $9.662^{\pm0.078}$ & - \\
      \midrule
      Axis-angle & \et{0.302}{0.004} & \et{0.464}{0.004} & \et{0.570}{0.004} & \et{3.702}{0.0855} & \et{4.807}{0.0160} & \et{8.452}{0.057} & \et{2.559}{0.078} \\
      Quaternion & \et{0.349}{0.005} & \et{0.533}{0.005} & $\underline{0.652^{\pm0.005}}$ & $\mathbf{0.748^{\pm0.0323}}$ & $\underline{4.293^{\pm0.0131}}$ & $\mathbf{9.797^{\pm0.064}}$ & $\mathbf{2.891^{\pm0.091}}$ \\
      Rotation matrix & \et{0.105}{0.001} & \et{0.193}{0.003} & \et{0.267}{0.002} & \et{91.074}{0.1451} & \et{10.190}{0.0120} & \et{3.385}{0.053} & \et{1.370}{0.037} \\
      6D rotation & $\mathbf{0.389^{\pm0.004}}$ & $\mathbf{0.577^{\pm0.005}}$ & $\mathbf{0.689^{\pm0.004}}$ & $\underline{1.195^{\pm0.0320}}$ & $\mathbf{3.963^{\pm0.0172}}$ & $\underline{9.906^{\pm0.075}}$ & \et{2.695}{0.090}\\
      Non-canonical & $\underline{0.353^{\pm0.004}}$ & $\underline{0.536^{\pm0.004}}$ & \et{0.646}{0.003} & \et{1.280}{0.0467} & \et{4.340}{0.0138} & \et{9.028}{0.073} & $\underline{2.855^{\pm0.100}}$ \\
    \bottomrule
  \end{tabular}}
\end{table*}

\begin{table}[t]
  \centering
  \caption{Quantitative results of interaction imitation on the Inter-X++ dataset. PHC* means that we fine-tune the PHC-Base~\cite{PHC} model on the Inter-X++ dataset with different epochs.}
  \label{tab:phc}
  \resizebox{0.48\textwidth}{!}{
  \begin{tabular}{lccccc}
    \toprule
    Method & $\text{Succ(\%)}$$\uparrow$ & $E_\text{g-mpjpe}$$\downarrow$ & $E_\text{mpjpe}$$\downarrow$ &  $\text{E}_{\text{acc}}$$\downarrow$ & $\text{E}_{\text{vel}}$$\downarrow$\\
    \midrule
      PHC-Base~\cite{PHC} & 76.79 & 70.26 & 70.19 & 14.51 & 13.34 \\
      \midrule
      PHC* (Epoch 6,000) & 80.17 & 61.68 & 61.97 & 11.92 & 11.87 \\
      PHC* (Epoch 7,500) & 80.21 & 65.29 & 63.64 & 12.33 & 12.27 \\
      PHC* (Epoch 9,000) & \textbf{81.30} & \textbf{58.58} & \textbf{61.95} & \textbf{11.72} & \textbf{11.75} \\
      PHC* (Epoch 10,500) & 81.20 & 59.73 & 62.76 & 12.11 & 12.06 \\
    \bottomrule
  \end{tabular}}
\end{table}

\subsubsection{Joint optimization and downstream transfer}
We jointly optimize the shared interaction encoder and the two complementary decoders using the overall objective
\begin{equation}
  \mathcal{L}_{\mathrm{all}}
  =
  \omega_{\mathrm{rec}}\mathcal{L}_{\mathrm{rec}}
  +
  \omega_{\mathrm{und}}\mathcal{L}_{\mathrm{und}},
  \label{eq:openhhi_loss}
\end{equation}
where $\omega_{\mathrm{rec}}$ and $\omega_{\mathrm{und}}$ balance the two objectives. During joint optimization, both losses are back-propagated through the shared interaction encoder, whereas each decoder is updated by its corresponding objective. In this manner, reconstruction prevents the representation from discarding fine-grained motion information, while caption generation encourages it to encode the action semantics conveyed by the textual descriptions. The resulting $\bm{h}$ therefore provides a common representation that supports both interaction generation and understanding.

Once jointly pretrained, OpenHHI serves as a shared interaction backbone for all downstream perceptive and generative tasks. The unified encoder maps interaction motions into a unified token space, eliminating the need to learn separate motion representations for different task families. For generative tasks, conditional models operate in this shared space and recover paired motions through the pretrained reconstruction and VQ-VAE decoders. For perceptive tasks, the same interaction tokens are supplied to lightweight classification, regression, or language heads. Only these newly introduced downstream modules need to be optimized. By sharing a single encoder and representation space across generative and perceptive tasks, OpenHHI transfers both fine-grained kinematic information and high-level interaction semantics, leading to consistent performance improvements across the downstream tasks.

\section{Experiments}

In this section, we conduct extensive experiments on four key components of Inter-X++: 1) Human-Human Interaction Representation, where we systematically compare and unify different interaction parameterizations; 2) Human-Human Interaction Imitation, where we evaluate physics-based motion regularization on Inter-X++; 3) Unified Interaction Model, where we benchmark OpenHHI across both generative and perceptive downstream tasks; and 4) Hierarchical Texts, where we investigate how textual descriptions at different levels of granularity affect interaction generation quality and motion-language alignment. We first describe the common experimental setup and evaluation protocol, followed by detailed quantitative and qualitative analyses of each component.

\subsection{Experimental Setup}

Following HumanML3D~\cite{humanml3d} and InterHuman~\cite{interhuman}, we split Inter-X++ into training, test, and validation sets with a ratio of 0.8, 0.15, and 0.05, respectively. While single-person motion sequences are typically canonicalized with respect to the first frame, we retain the global translations of both participants to preserve their relative spatial configuration. For diffusion-based models~\cite{ddpm}, we employ 1,000 diffusion steps during training and a five-step DDIM sampler~\cite{ddim} during inference. All models are trained using NVIDIA A100 GPUs.

For the \textbf{interaction representation} study, we instantiate the same text-conditioned interaction generation model with the five motion representations introduced in~\cref{sec:hhi_rep}, including axis-angle, quaternion, rotation matrix, 6D rotation, and the non-canonical representation of InterGen~\cite{interhuman}. To isolate the effect of the representation, we keep the network architecture, data split, training schedule, and all other configurations identical across variants. Since their outputs lie in different feature spaces, all generated outputs are projected into a unified 6D rotation space~\cite{actor} for standardized evaluation. For \textbf{interaction imitation}, we build upon the implementation of PHC~\cite{PHC}. We directly evaluate the original PHC-Base policy trained on AMASS as the baseline and further train it on individual Inter-X++ motion sequences in a single-agent simulation environment, following MultiPhys~\cite{ugrinovic2024multiphys}. At inference time, two independently controlled humanoid agents are placed in a shared simulation environment, where contact and collision constraints are jointly enforced to reduce foot sliding, motion jittering, and inter-person penetration. For the \textbf{unified interaction model}, we establish a comprehensive benchmark covering all eight generative and perceptive downstream tasks, evaluate representative baseline methods, and further conduct experiments using our proposed shared OpenHHI encoder. For the \textbf{hierarchical text} study, we evaluate text-conditioned HHI generation using descriptions at different levels of granularity.

\subsection{Evaluation Protocol}

We summarize the adopted evaluation metrics according to task type. For text-conditioned HHI generation, interaction representation, and hierarchical text experiments, we follow HumanML3D~\cite{humanml3d} and InterHuman~\cite{interhuman} and report Top-1/2/3 R-Precision, Frechet Inception Distance (FID)~\cite{fid}, MultiModal Distance (MM Dist), Diversity, and MModality. R-Precision and MM Dist evaluate motion-language alignment, FID measures the distributional discrepancy from real interactions, and Diversity and MModality characterize overall and condition-specific motion variation, respectively. For action-conditioned HHI generation, human reaction generation, and stylized HHI generation, we report FID, action recognition accuracy, Diversity, and MModality, where the accuracy is computed using a ST-GCN~\cite{stgcn} evaluator to measure condition consistency. Following PHC~\cite{PHC}, interaction imitation is evaluated using the success rate (Succ), global and root-relative mean per-joint position errors ($E_\text{g-mpjpe}$ and $E_\text{mpjpe}$), acceleration error ($E_\text{acc}$), and velocity error ($E_\text{vel}$). For perceptive tasks, HHI recognition uses Top-1 and Top-5 accuracy, causal interaction order inference uses binary classification accuracy, HHI captioning uses BLEU~\cite{bleu}, ROUGE~\cite{rouge}, CIDEr~\cite{cider}, and BERTScore~\cite{bertscore}, and personality assessment reports the coefficient of determination ($R^2$) for each personality trait. Higher R-Precision, classification accuracy, captioning scores, $R^2$, and Succ indicate better performance, whereas lower FID, MM Dist, and imitation errors are preferred; for Diversity and MModality, $\rightarrow$ denotes proximity to the real-data statistics and $\uparrow$ denotes greater variation. Unless otherwise specified, stochastic generation evaluations are repeated 20 times and reported with 95\% confidence intervals, while MModality for text-conditioned generation is evaluated five times.

\begin{table*}[t]
  \centering
  \caption{Experimental results of text-conditioned HHI generation on the Inter-X++ dataset, where $\pm$ indicates 95\% confidence interval and $\rightarrow$ means the closer the better. * denotes our reproduced results based on the official codebase.}
  \label{tab:t2m_sota}
  \resizebox{1\textwidth}{!}{
  \begin{tabular}{lccccccc}
    \toprule
    \multirow{2}{*}{Method}  & \multicolumn{3}{c}{R Precision$\uparrow$} & \multirow{2}{*}{FID $\downarrow$} & \multirow{2}{*}{MM Dist$\downarrow$}  & \multirow{2}{*}{Diversity$\rightarrow $} & \multirow{2}{*}{MModality $\uparrow$}\\
    \cmidrule(lr){2-4} & Top 1 & Top 2  & Top 3 \\
    \midrule
      Real & $0.448^{\pm0.003}$ & $0.639^{\pm0.003}$ & $0.743^{\pm0.003}$ & $0.001^{\pm0.0001}$ & $3.652^{\pm0.009}$ & $9.662^{\pm0.078}$ & - \\
      \midrule
      TEMOS~\cite{petrovich22temos} & $0.092^{\pm0.003}$ & $0.171^{\pm0.003}$ & $0.238^{\pm0.002}$ & $29.258^{\pm0.0694}$ & $6.867^{\pm0.013}$ & $4.738^{\pm0.078}$ & $0.672^{\pm0.041}$ \\ 
      T2M~\cite{humanml3d} & $0.184^{\pm0.010}$ & $0.298^{\pm0.006}$ & $0.396^{\pm0.005}$ & $5.481^{\pm0.3820}$ & $9.576^{\pm 0.006}$ & $5.771^{\pm0.151}$ & $2.761^{\pm 0.042}$\\
      MDM~\cite{mdm} &  $0.203^{\pm0.009}$ & $0.329^{\pm0.007}$ & $0.426^{\pm0.005}$ & $23.701^{\pm0.0569}$ & $9.548^{\pm 0.014}$ & $5.856^{\pm0.077}$ & $\underline{3.490^{\pm0.061}}$\\
      MDM(GRU)~\cite{mdm} &  $0.179^{\pm0.006}$ & $0.299^{\pm0.005}$ & $0.387^{\pm0.007}$ & $32.617^{\pm0.1221}$ & $9.557^{\pm0.019}$ & $7.003^{\pm0.134}$ & $3.430^{\pm0.035}$\\
      ComMDM~\cite{commdm} & $0.090^{\pm0.002}$ & $0.165^{\pm0.004}$ & $0.236^{\pm0.004}$ & $29.266^{\pm0.0668}$ & $6.870^{\pm 0.017}$ & $4.734^{\pm0.067}$ & $0.771^{\pm 0.053}$\\
      InterGen~\cite{interhuman} & $0.207^{\pm0.004}$ & $0.335^{\pm0.005}$ & $0.429^{\pm0.005}$ & $5.207^{\pm0.2160}$ & $9.580^{\pm 0.011}$ & $7.788^{\pm0.208}$ & $\mathbf{3.686^{\pm 0.052}}$\\
      InterMask*~\cite{javed2024intermask} & $\underline{0.389^{\pm0.004}}$ & $\underline{0.577^{\pm0.005}}$ & $\underline{0.689^{\pm0.004}}$ & $\underline{1.195^{\pm0.0320}}$ & $\underline{3.963^{\pm0.0172}}$ & $\underline{9.906^{\pm0.075}}$ & $2.695^{\pm0.090}$\\
      OpenHHI (Ours) & $\mathbf{0.437^{\pm0.005}}$ & $\mathbf{0.627^{\pm0.004}}$ & $\mathbf{0.732^{\pm0.005}}$ & $\mathbf{0.573^{\pm0.0209}}$ & $\mathbf{3.599^{\pm0.017}}$ & $\mathbf{9.685^{\pm0.085}}$ & $1.796^{\pm0.071}$ \\
    \bottomrule
  \end{tabular}}
\end{table*}

\begin{figure*}[t]
  \centering
   \includegraphics[width=1.0\linewidth]{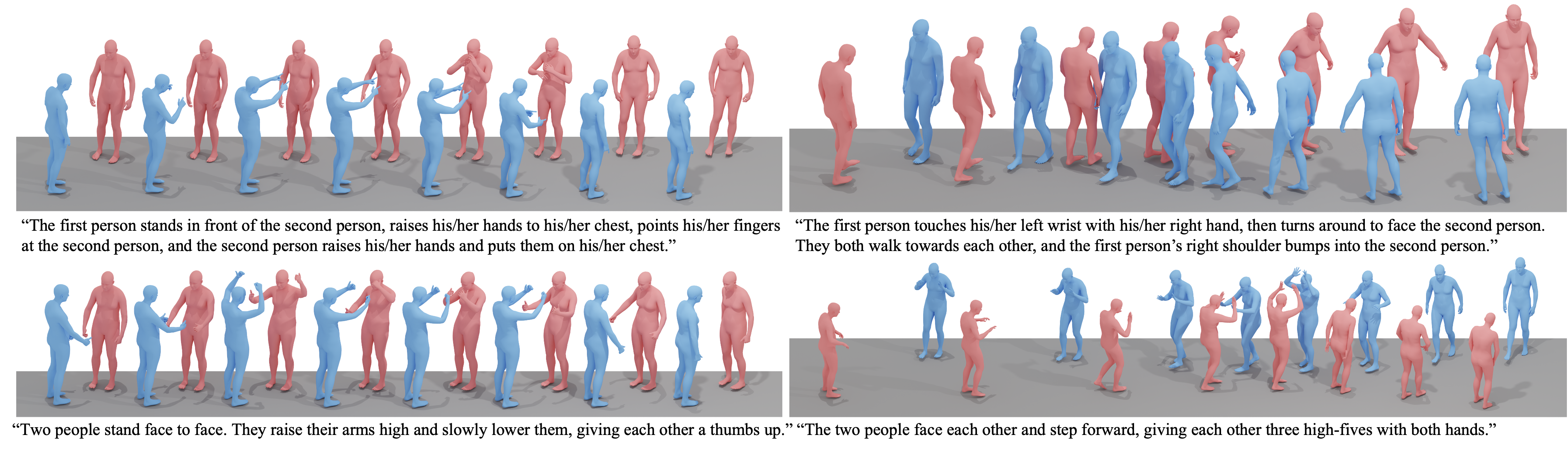}
   \caption{\textbf{Visualization results} of text-conditioned human-human interaction generation on the Inter-X++ dataset. Text is provided as the input condition, yielding interaction sequences as the generated output.}
   \label{fig:vis}
\end{figure*}

\subsection{Human-Human Interaction Representation}

To systematically investigate the effect of HHI representations, we train an identical text-conditioned HHI generation model using axis-angle, quaternion, rotation matrix, 6D rotation, and non-canonical representations, strictly keeping the network architecture, data splits, training schedule, and evaluation protocol unchanged. Because these parameterizations reside in distinct feature spaces, all generated outputs are projected into a unified 6D rotation space for standardized evaluation. As reported in~\cref{tab:ablation_data_rep}, the 6D rotation representation achieves the best overall performance, attaining the highest Top-1, Top-2, and Top-3 R-Precision scores alongside the lowest MM-Dist. Although the quaternion representation yields superior FID and MModality scores, its retrieval accuracy and cross-modal semantic alignment remain consistently inferior to those of the 6D rotation. Furthermore, while the non-canonical representation shows competitive potential, it remains less effective overall, and both axis-angle and rotation matrix parameterizations suffer from substantial performance degradation. These empirical results demonstrate that the continuous and compact nature of the 6D parameterization strikes an optimal balance among optimization stability, generation fidelity, and semantic consistency. Consequently, we adopt the 6D rotation representation across all subsequent experiments.

\subsection{Human-Human Interaction Imitation}

The results in~\cref{tab:phc} highlight the benefit of adapting the imitation policy to interaction data. The limited transferability of PHC-Base indicates that motion priors learned from individual human motions are insufficient for reproducing the coordinated dynamics of two-person interactions. In-domain optimization consistently raises the success rate while reducing both spatial tracking errors and temporal motion errors. Among the evaluated checkpoints, the model trained for 9,000 epochs ranks first across every metric, demonstrating more accurate global and root-relative tracking together with smoother motion reproduction. Extending training to 10,500 epochs yields a slight decline rather than further gains, suggesting that performance has already saturated. We therefore select the 9,000 epoch checkpoint for physics-based refinement of the interaction motions in Inter-X++.

\begin{table}[t]
  \centering
  \caption{Experimental results of action-conditioned interaction generation on the Inter-X++ dataset.}
  \label{tab:a2m_sota}
  \resizebox{0.48\textwidth}{!}{
  \begin{tabular}{lcccc}
    \toprule
    Method & FID$\downarrow$ & Acc.$\uparrow$ & Div.$\rightarrow$ & Multimod.$\rightarrow$\\
    \midrule
      Real & $0.281^{\pm0.002}$ & $0.990^{\pm0.0000}$ & $12.890^{\pm0.028}$ & $22.391^{\pm0.195}$ \\
      \midrule
      Action2Motion~\cite{action2motion} & $20.295^{\pm12.081}$ & $0.766^{\pm0.0003}$ & $11.581^{\pm0.024}$ & $15.345^{\pm0.245}$\\
      ACTOR~\cite{actor} & $9.392^{\pm0.816}$ & $0.855^{\pm0.0003}$ & $11.594^{\pm0.029}$ & $15.327^{\pm0.195}$ \\
      MDM~\cite{mdm} & $12.426^{\pm2.584}$ & $0.896^{\pm0.0004}$ & $13.492^{\pm0.033}$ & $\mathbf{22.042^{\pm0.153}}$\\
      MDM(GRU)~\cite{mdm} & $35.003^{\pm7.876}$ & $0.716^{\pm0.0006}$ & $\underline{12.579^{\pm0.038}}$ & $16.456^{\pm0.100}$\\
      Actformer~\cite{xu2023actformer} & $\underline{8.067^{\pm0.653}}$ & $\underline{0.945^{\pm0.0007}}$ & $12.512^{\pm0.05}$ & $16.187^{\pm0.189}$ \\
      OpenHHI (Ours) & $\mathbf{4.215^{\pm0.284}}$ & $\mathbf{0.969^{\pm0.0004}}$ & $\mathbf{12.873^{\pm0.035}}$ & $\underline{17.184^{\pm0.142}}$ \\
    \bottomrule
  \end{tabular}}
\end{table}

\begin{table}[t]
  \centering
  \caption{Experimental results of human reaction generation based on action labels on the Inter-X++ dataset.}
  \label{tab:regen_sota}
  \resizebox{0.48\textwidth}{!}{
  \begin{tabular}{lcccc}
    \toprule
    Method & FID$\downarrow$ & Acc.$\uparrow$ & Div.$\rightarrow$ & Multimod.$\rightarrow$\\
    \midrule
      Real & $0.260^{\pm0.0021}$ & $0.988^{\pm0.0000}$ & $12.115^{\pm0.031}$ & $21.498^{\pm0.131}$ \\
      \midrule
      MDM~\cite{mdm} & $6.747^{\pm0.3153}$ & $0.903^{\pm0.0001}$ & $12.264^{\pm0.051}$ & $19.681^{\pm0.234}$ \\
      MDM(GRU)~\cite{mdm} & $19.968^{\pm1.1700}$ & $0.752^{\pm0.0003}$ & $12.351^{\pm0.049}$ & $18.056^{\pm0.156}$\\
      RAIG~\cite{role_aware} & $6.372^{\pm0.2154}$ & $0.908^{\pm0.0001}$ & $12.330^{\pm0.060}$ & $\underline{20.071^{\pm0.299}}$ \\
      AGRoL~\cite{agrol} & $\underline{4.386^{\pm0.2186}}$ & $\underline{0.925^{\pm0.0001}}$ & $\underline{12.204^{\pm0.042}}$ & $\mathbf{20.199^{\pm0.226}}$\\
      OpenHHI (Ours) & $\mathbf{3.172^{\pm0.1645}}$ & $\mathbf{0.947^{\pm0.0001}}$ & $\mathbf{12.132^{\pm0.039}}$ & $18.873^{\pm0.185}$ \\
    \bottomrule
  \end{tabular}}
\end{table}

\subsection{Unified Human-Human Interaction Model}

In this section, we first establish a unified benchmark for the proposed eight downstream tasks and provide representative methods as baselines under a consistent evaluation protocol. We then apply OpenHHI to every task as a shared HHI encoder and interaction representation, introducing only lightweight downstream modules for different prediction objectives. This comprehensive evaluation examines whether a single unified representation can transfer consistently across diverse conditioning modalities and both generative and perceptive tasks.

\subsubsection{Text-conditioned Interaction Generation}

We compare seven representative text-to-motion and interaction generation methods: TEMOS~\cite{petrovich22temos}, T2M~\cite{humanml3d}, MDM~\cite{mdm}, MDM-GRU~\cite{mdm}, ComMDM~\cite{commdm}, InterGen~\cite{interhuman}, and InterMask~\cite{javed2024intermask}. The single-person models are extended to synthesize paired motions, and all methods are evaluated under the unified protocol. As shown in~\cref{tab:t2m_sota}, InterMask yields the best results among the compared baselines for all three R-Precision scores, FID, MM Dist, and Diversity, whereas InterGen produces the highest MModality. The substantial gaps between earlier baselines and InterMask also show that jointly modeling the two participants is essential for text-conditioned HHI generation. OpenHHI further delivers substantial overall improvements over these competing baselines across the principal retrieval and distributional metrics. The consistent gains demonstrate that its unified HHI representation supports more accurate text-interaction alignment and higher-quality motion generation while retaining realistic motion diversity.

The visualization examples in~\cref{fig:vis} illustrate that models trained on Inter-X++ can express detailed hand movements and coordinated two-person dynamics. For interactions such as ``Wave'', ``High five'' and ``Thumb up'', the generated sequences exhibit clearer limb articulation and plausible relative positioning than the counterparts trained on InterHuman, whose lack of dexterous hand annotations leads to ambiguous gestures and more visible penetration artifacts. Additional visual comparisons and video results are provided in the supplementary.

\begin{table}[t]
  \centering
  \caption{Experimental results of skeleton-based human-human interaction recognition on the Inter-X++ dataset.}
  \label{tab:m2a_sota}
  \begin{tabular}{lcc}
    \toprule
    Method & Top-1 (\%) & Top-5 (\%) \\
    \midrule
      ST-GCN~\cite{stgcn} & 64.62 & 90.16\\
      2s-AGCN~\cite{2s-agcn} & 75.22 & 93.73 \\
      HD-GCN~\cite{hdgcn} & 77.40 & 94.73\\
      CTR-GCN~\cite{ctrgcn} & 82.19 & 96.72\\
      MS-G3D~\cite{ms-g3d} & \underline{83.30} & \underline{97.09}\\
      OpenHHI (Ours) & \textbf{84.88} & \textbf{97.13}\\
    \bottomrule
  \end{tabular}
\end{table}

\begin{table}[t]
  \centering
  \caption{Experimental results of human-human interaction captioning on the Inter-X++ dataset.}
  \label{tab:m2t_sota}
  \resizebox{0.48\textwidth}{!}{
  \begin{tabular}{lccccc}
    \toprule
    Method & Bleu@1$\uparrow$ & Bleu@4 $\uparrow$ & Rouge $\uparrow$ & Cider $\uparrow$ & BertScore $\uparrow$\\
    \midrule
      RAEs~\cite{yamada2018paired} & 28.6 & 9.7 & 34.1 & 25.9 & 10.2 \\
      Seq2Seq~\cite{plappert2018learning} & 53.8 & 18.5 & 45.2 & \underline{61.9} & 27.1 \\
      SeqGAN~\cite{goutsu2021linguistic} & 45.4 & 14.1 & 36.8 & 52.3 & 21.4 \\
      TM2T~\cite{guo2022tm2t} & \underline{56.8} & \underline{21.6} & \underline{48.2} & \textbf{75.5} & \underline{32.7}\\
      OpenHHI (Ours) & \textbf{70.7} & \textbf{35.4} & \textbf{48.9} & 49.1 & \textbf{35.1} \\
    \bottomrule
  \end{tabular}}
\end{table}

\subsubsection{Action-conditioned Interaction Generation}

Inter-X++ provides 40 semantic categories for action-conditioned HHI generation. We adapt Action2Motion~\cite{action2motion}, ACTOR~\cite{actor}, MDM~\cite{mdm}, MDM-GRU~\cite{mdm}, and Actformer~\cite{xu2023actformer} to human-human interactions and evaluate them using the same data split as in text-conditioned generation. As shown in~\cref{tab:a2m_sota}, Actformer obtains the lowest FID and highest recognition accuracy among the baselines, whereas MDM-GRU and MDM more closely match different real-motion variation statistics. This split outcome highlights the difficulty of balancing motion fidelity, condition consistency, and generation diversity. OpenHHI delivers substantial overall improvements over these existing baselines, particularly in distribution fidelity and action consistency, while preserving motion variation close to the real data. These gains show that the unified HHI representation transfers effectively to discrete action control.

\subsubsection{Human Reaction Generation}

Human reaction generation predicts the reactor's motion from the observed motion of the actor. We evaluate adapted versions of MDM~\cite{mdm} and MDM-GRU~\cite{mdm} together with the RAIG~\cite{role_aware} and AGRoL~\cite{agrol}. Among the baselines as shown in~\cref{tab:regen_sota}, AGRoL achieves the best results across the evaluation metrics, confirming the value of explicitly modeling the dependency between the initiating and reactive motions. OpenHHI consistently improves upon the baselines across all evaluation criteria. The considerable gains indicate that its unified HHI representation effectively captures the coupling between an initiating motion and its corresponding response, producing reactions that are both condition-consistent and diverse.

\begin{figure*}[t]
  \centering
   \includegraphics[width=1.0\linewidth]{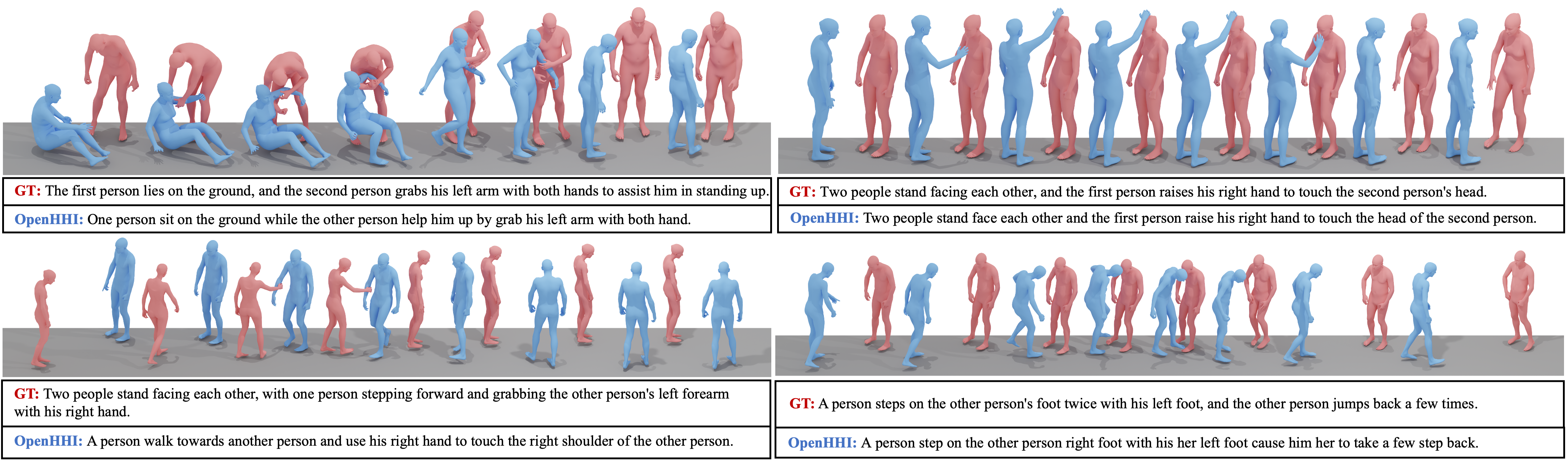}
   \caption{Qualitative results of human-human interaction captioning on Inter-X++. Each example presents a sampled interaction sequence together with its ground-truth description and the caption generated by OpenHHI. The predictions coherently describe the coordinated behaviors of both participants, including their roles, body movements, and interaction details.}
   \label{fig:caption_vis}
\end{figure*}

\begin{table}[t]
  \centering
  \caption{Experimental results of causal interaction order inference on the Inter-X++ dataset.}
  \label{tab:order_ass}
  \begin{tabular}{lc}
    \toprule
    Method & Accuracy (\%)\\
    \midrule
      ST-GCN~\cite{stgcn} & 62.3\\
      2s-AGCN~\cite{2s-agcn} & 68.2 \\
      HD-GCN~\cite{hdgcn} & 70.6\\
      CTR-GCN~\cite{ctrgcn} & 74.5\\
      MS-G3D~\cite{ms-g3d} & \underline{76.8}\\
      OpenHHI (Ours) & \textbf{79.1}\\
    \bottomrule
  \end{tabular}
\end{table}

\subsubsection{Human Interaction Recognition}
We evaluate ST-GCN~\cite{stgcn}, 2s-AGCN~\cite{2s-agcn}, HD-GCN~\cite{hdgcn}, CTR-GCN~\cite{ctrgcn}, and MS-G3D~\cite{ms-g3d} using only the skeleton-joint stream, without ensembling bone or motion streams. As shown in~\cref{tab:m2a_sota}, MS-G3D obtains the best Top-1 and Top-5 accuracy among the compared baselines. The remaining recognition errors indicate that distinguishing interactions with similar global trajectories requires sensitivity to fine-grained gestures and actor-reactor variations. OpenHHI further improves interaction recognition performance over these traditional skeleton-based baselines. The gains demonstrate that joint reconstruction and language supervision produce a unified HHI representation that is more discriminative for closely related interaction categories.

\subsubsection{Human interaction captioning}

Human interaction captioning maps paired motion sequences to natural language descriptions. Following the text-generation split, we adapt RAEs~\cite{yamada2018paired}, Seq2Seq~\cite{plappert2018learning}, SeqGAN~\cite{goutsu2021linguistic}, and TM2T~\cite{guo2022tm2t} to two-person inputs. As depicted in~\cref{tab:m2t_sota}, TM2T provides the best baseline results across the language evaluation metrics, while the recurrent autoencoder performs substantially worse, reflecting its difficulty in retaining long-range interaction dynamics. The attention-based and adversarial alternatives improve over RAEs but still leave considerable room for fine-grained interaction description. OpenHHI achieves the best BLEU@1, BLEU@4, ROUGE, and BERTScore results, demonstrating stronger lexical overlap and semantic similarity. However, its CIDEr score of 49.1 remains below the 75.5 achieved by TM2T, indicating that further improvements are needed in consensus-oriented caption generation. The qualitative examples in~\cref{fig:caption_vis} further demonstrate that the generated captions are semantically coherent and accurately reflect the fine-grained interaction details between the two participants.

\begin{table}[t]
  \centering
  \caption{Experimental results of action-conditioned stylized human interaction generation on the Inter-X++ dataset.}
  \label{tab:style_gen}
  \resizebox{0.48\textwidth}{!}{
  \begin{tabular}{lcccc}
    \toprule
    Method & FID$\downarrow$ & Acc.$\uparrow$ & Div.$\rightarrow$ & Multimod.$\rightarrow$\\
    \midrule
      Real & $0.281^{\pm0.002}$ & $0.990^{\pm0.0000}$ & $12.890^{\pm0.028}$ & $22.391^{\pm0.195}$ \\
      \midrule
      Action2Motion~\cite{action2motion} & $21.182^{\pm13.319}$ & $0.737^{\pm0.0005}$ & $11.492^{\pm0.032}$ & $14.934^{\pm0.258}$ \\
      ACTOR~\cite{actor} & $9.796^{\pm0.862}$ & $0.867^{\pm0.0003}$ & $11.862^{\pm0.039}$ & $15.174^{\pm0.245}$\\
      MDM~\cite{mdm} & $11.762^{\pm1.854}$ & $0.912^{\pm0.0002}$ & $\underline{13.025^{\pm0.028}}$ & $\mathbf{21.742^{\pm0.106}}$\\
      MDM(GRU)~\cite{mdm} & $31.688^{\pm4.492}$ & $0.753^{\pm0.0006}$ & $12.259^{\pm0.039}$ & $16.271^{\pm0.206}$\\
      Actformer~\cite{xu2023actformer} & $\underline{8.544^{\pm0.684}}$ & $\underline{0.932^{\pm0.0006}}$ & $12.116^{\pm0.062}$ & $16.122^{\pm0.183}$\\
      OpenHHI (Ours) & $\mathbf{5.286^{\pm0.421}}$ & $\mathbf{0.951^{\pm0.0004}}$ & $\mathbf{12.914^{\pm0.044}}$ & $\underline{17.018^{\pm0.162}}$ \\
    \bottomrule
  \end{tabular}}
\end{table}

\begin{table}[t]
  \centering
  \caption{The $R^2$ results (\%) of the personality assessment on the Inter-X++ dataset.}
  \label{tab:per_ass}
  \resizebox{0.48\textwidth}{!}{
  \begin{tabular}{lccccc}
    \toprule
    Method & Openness & Conscientiousness & Extraversion & Agreeableness & Neuroticism \\
    \midrule
      ST-GCN~\cite{stgcn} & 21.16 & 25.38 & 34.91 & 23.67 & 13.02 \\
      2s-AGCN~\cite{2s-agcn} & 23.46 & 31.27 & 38.72 & 24.88 & 13.57 \\
      HD-GCN~\cite{hdgcn} & 25.92 & 33.19 & 41.33 & 26.83 & 14.29\\
      CTR-GCN~\cite{ctrgcn} & 27.78 & 35.41 & 43.52 & \underline{29.43} & 15.63\\
      MS-G3D~\cite{ms-g3d} & \underline{28.36} & \underline{37.88} & \underline{46.23} & 29.07 & \underline{16.35}\\
      OpenHHI (Ours) & \textbf{30.74} & \textbf{40.12} & \textbf{48.67} & \textbf{31.05} & \textbf{18.21} \\
    \bottomrule
  \end{tabular}}
\end{table}

\subsubsection{Causal order inference}

Causal order inference determines which participant initiates an interaction and which participant responds. We formulate it as a binary classification and evaluate the same five graph-based backbones used for interaction recognition. As shown in~\cref{tab:order_ass}, MS-G3D obtains the best accuracy among these baselines. OpenHHI surpasses the compared methods and yields a clear improvement in causal order inference. This result suggests that the unified HHI representation retains directional interaction cues and participant-specific dynamics that are essential for identifying the actor and reactor.

\subsubsection{Stylized human interaction generation}
For stylized generation, we augment Action2Motion~\cite{action2motion}, ACTOR~\cite{actor}, MDM~\cite{mdm}, MDM-GRU~\cite{mdm}, and Actformer~\cite{xu2023actformer} with the relationship familiarity level as a style code. Among the baselines in~\cref{tab:style_gen}, Actformer achieves the lowest FID and highest recognition accuracy, whereas MDM more closely approaches the real-data motion variation statistics. The outcome suggests that maintaining interaction semantics while expressing relationship-dependent styles remains challenging for the evaluated generators. OpenHHI achieves substantial improvements in both distribution fidelity and condition consistency while preserving realistic motion variation. This balanced performance indicates that the unified HHI representation retains the underlying interaction category while allowing relationship cues to modulate how an interaction is performed.

\subsubsection{Personality assessment}
Personality assessment predicts five personality traits from interaction motions. To prevent identity leakage, we split the training, validation, and test data by participant identity and formulate each trait as a regression target. Among the graph-based baselines in~\cref{tab:per_ass}, MS-G3D performs best on most personality dimensions, while CTR-GCN gives the highest Agreeableness score. OpenHHI consistently improves the prediction results across all five personality dimensions. These gains suggest that its unified HHI representation encodes not only categorical interaction semantics but also subtle behavioral patterns associated with individual motion style. However, the comparatively low $R^2$ values confirm that personality cues are considerably subtler and more challenging than categorical action semantics.

\begin{table*}[t]
  \centering
  \caption{Ablation of the relative weighting between the reconstruction and interaction-understanding objectives in OpenHHI for Level-3 text-conditioned HHI generation. We fix $\omega_{\mathrm{rec}}=1$ and vary $\rho=\omega_{\mathrm{und}}/\omega_{\mathrm{rec}}$. InterMask* denotes our reproduced baseline using the official implementation. Bold and underlined values indicate the best and second-best results, respectively.}
  \label{tab:ablation_weight}
  \resizebox{1\textwidth}{!}{
  \begin{tabular}{lccccccc}
    \toprule
    \multirow{2}{*}{$\rho$}  & \multicolumn{3}{c}{R Precision$\uparrow$} & \multirow{2}{*}{FID $\downarrow$} & \multirow{2}{*}{MM Dist$\downarrow$}  & \multirow{2}{*}{Diversity$\rightarrow $} & \multirow{2}{*}{MModality $\uparrow$}\\
    \cmidrule(lr){2-4} & Top 1 & Top 2  & Top 3 \\
    \midrule
      Real & $0.448^{\pm0.003}$ & $0.639^{\pm0.003}$ & $0.743^{\pm0.003}$ & $0.001^{\pm0.0001}$ & $3.652^{\pm0.009}$ & $9.662^{\pm0.078}$ & - \\
      InterMask*~\cite{javed2024intermask} & $0.389^{\pm0.004}$ & $0.577^{\pm0.005}$ & $0.689^{\pm0.004}$ & $1.195^{\pm0.0320}$ & $3.963^{\pm0.0172}$ & $9.906^{\pm0.075}$ & $\mathbf{2.695^{\pm0.090}}$\\
      \midrule
      $0.1$ & $0.422^{\pm0.005}$ & $0.613^{\pm0.004}$ & $0.722^{\pm0.004}$ & $\mathbf{0.366^{\pm0.0183}}$ & $3.660^{\pm0.018}$ & $9.616^{\pm0.085}$ & $\underline{1.901^{\pm0.057}}$ \\
      $0.5$ & $\mathbf{0.437^{\pm0.005}}$ & $\mathbf{0.627^{\pm0.004}}$ & $\underline{0.732^{\pm0.005}}$ & $0.573^{\pm0.0209}$ & $\mathbf{3.599^{\pm0.017}}$ & $\mathbf{9.685^{\pm0.085}}$ & $1.796^{\pm0.071}$ \\
      $1.0$ & $\underline{0.429^{\pm0.005}}$ & $\underline{0.624^{\pm0.005}}$ & $\mathbf{0.733^{\pm0.005}}$ & $\underline{0.480^{\pm0.0196}}$ & $\underline{3.602^{\pm0.021}}$ & $9.748^{\pm0.084}$ & $1.784^{\pm0.050}$ \\
      $2.0$ & $0.407^{\pm0.004}$ & $0.596^{\pm0.004}$ & $0.705^{\pm0.003}$ & $0.998^{\pm0.0279}$ & $3.805^{\pm0.013}$ & $\underline{9.693^{\pm0.101}}$ & $1.829^{\pm0.060}$ \\
    \bottomrule
  \end{tabular}}
\end{table*}

\begin{table*}[t]
  \centering
  \caption{Quantitative results of OpenHHI for text-conditioned HHI generation using hierarchical descriptions of different granularities. ``Real'' reports the statistics of ground-truth motions paired with the corresponding text level.}
  \label{tab:hierarchical_text}
  \resizebox{1\textwidth}{!}{
  \begin{tabular}{llccccccc}
    \toprule
    \multirow{2}{*}{Texts} & \multirow{2}{*}{Method}  & \multicolumn{3}{c}{R Precision$\uparrow$} & \multirow{2}{*}{FID $\downarrow$} & \multirow{2}{*}{MM Dist$\downarrow$}  & \multirow{2}{*}{Diversity$\rightarrow $} & \multirow{2}{*}{MModality $\uparrow$}\\
    \cmidrule(lr){3-5} & & Top 1 & Top 2  & Top 3 \\
    \midrule
      \multirow{2}{*}{Level-1} & Real & $0.394^{\pm0.003}$ & $0.584^{\pm0.004}$ & $0.694^{\pm0.003}$ & $0.001^{\pm0.0001}$ & $3.719^{\pm0.011}$ & $9.013^{\pm0.072}$ & - \\
      & OpenHHI & $\mathbf{0.376^{\pm0.004}}$ & $\mathbf{0.549^{\pm0.004}}$ & $\mathbf{0.647^{\pm0.003}}$ & $\mathbf{2.654^{\pm0.0516}}$ & $\mathbf{4.187^{\pm0.018}}$ & $\mathbf{8.060^{\pm0.070}}$ & $2.086^{\pm0.074}$ \\
      \midrule
      \multirow{2}{*}{Level-2} & Real & $0.398^{\pm0.004}$ & $0.591^{\pm0.004}$ & $0.699^{\pm0.003}$ & $0.001^{\pm0.0001}$ & $3.704^{\pm0.011}$ & $8.819^{\pm0.076}$ & - \\
      & OpenHHI & $\mathbf{0.384^{\pm0.004}}$ & $\mathbf{0.566^{\pm0.004}}$ & $\mathbf{0.669^{\pm0.004}}$ & $\mathbf{2.774^{\pm0.0625}}$ & $\mathbf{4.053^{\pm0.015}}$ & $\mathbf{8.401^{\pm0.076}}$ & $2.036^{\pm0.057}$ \\ 
      \midrule
      \multirow{2}{*}{Level-3} & Real & $0.448^{\pm0.003}$ & $0.639^{\pm0.003}$ & $0.743^{\pm0.003}$ & $0.001^{\pm0.0001}$ & $3.652^{\pm0.009}$ & $9.662^{\pm0.078}$ & - \\
      & OpenHHI & $\mathbf{0.437^{\pm0.005}}$ & $\mathbf{0.627^{\pm0.004}}$ & $\mathbf{0.732^{\pm0.005}}$ & $\mathbf{0.573^{\pm0.0209}}$ & $\mathbf{3.599^{\pm0.017}}$ & $\mathbf{9.685^{\pm0.085}}$ & $1.796^{\pm0.071}$ \\
    \bottomrule
  \end{tabular}}
\end{table*}

\subsubsection{Balancing Reconstruction and Understanding Objectives}

We study the balance between the interaction reconstruction and understanding objectives in~\cref{eq:openhhi_loss} by fixing $\omega_{\mathrm{rec}}=1$ and varying $\rho=\omega_{\mathrm{und}}/\omega_{\mathrm{rec}}$ over $\{0.1,0.5,1.0,2.0\}$. As shown in~\cref{tab:ablation_weight}, a small $\rho$ favors motion fidelity but weakens motion-language alignment, whereas an overly large $\rho$ degrades distributional quality by suppressing motion details. The setting $\rho=0.5$ provides the best overall balance across R-Precision, MM Dist, and Diversity, confirming the complementary roles of the two objectives; we therefore adopt it as the default for all the downstream tasks.

\subsection{Hierarchical Texts for Generation}

\begin{figure*}[t]
  \centering
   \includegraphics[width=1.0\linewidth]{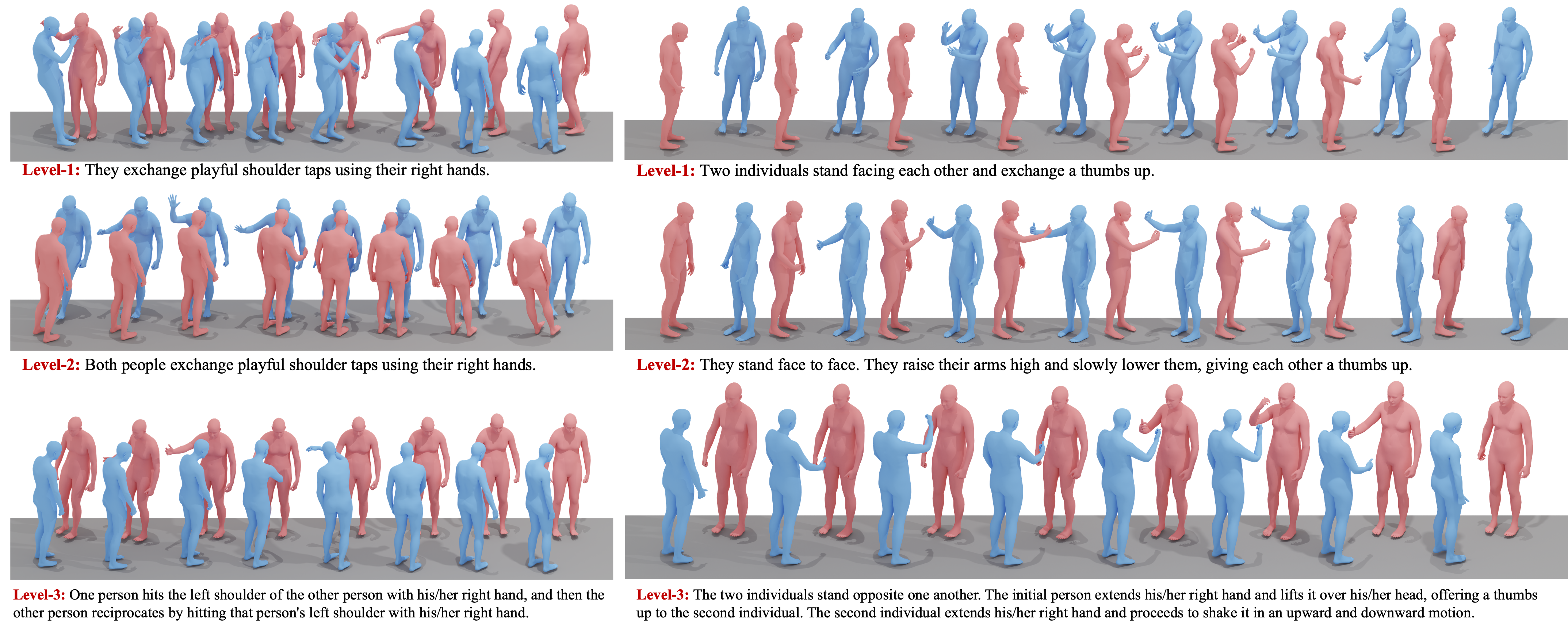}
   \caption{Qualitative comparison of HHI generation conditioned on hierarchical descriptions from Inter-X++. Each column presents motions generated for the same interaction at three levels of textual granularity. Coarser prompts permit broader motion variations, whereas finer descriptions provide more precise control.}
   \label{fig:level3_vis}
\end{figure*}

We examine how textual granularity affects HHI generation by evaluating OpenHHI with three description levels under identical settings. Level-1 uses short descriptions simplified from the original annotations, Level-2 provides more complete summaries of the overall interaction, and Level-3 specifies participant movements, body-part actions, and spatial relations. This comparison isolates the effect of increasingly detailed language conditions. 
As shown in~\cref{tab:hierarchical_text}, Level-3 achieves the highest R-Precision, indicating that fine-grained descriptions enable the model to learn more accurate motion-language alignment. In contrast, shorter descriptions omit participant-specific and spatial details, increasing the ambiguity between a prompt and its valid motion realizations. The qualitative results in~\cref{fig:level3_vis} further show that detailed prompts provide stronger control over individual actions and relative positioning, whereas simpler prompts impose fewer constraints and therefore allow greater generation diversity.

\section{Conclusion and Limitations}

Inter-X++ brings accurately captured whole-body and hand motions, physics-refined interaction sequences, and a broad annotation suite into one resource for HHI research. Its hierarchical narratives, action semantics, causal role order, contact cues, and social attributes expose complementary aspects of how two people move, respond, and relate to one another, supporting evaluation across versatile downstream tasks. Our extensive experiments examine this resource from multiple angles: the interaction representation study establishes a consistent basis for comparison, physical imitation validates the benefit of interaction-domain adaptation, and hierarchical-text evaluation reveals how linguistic detail trades generation diversity for controllability. Results throughout the benchmark further show that OpenHHI can reuse one interaction encoder across varied objectives while retaining both kinematic and semantic information. Together, our data, annotations, unified representation and systematic evaluation provide a solid basis for promoting in-depth research works on HHI analysis.

\noindent{\textbf{Limitations}.}
Our work has some limitations in the following aspects. 1) \textbf{Facial expressions.} Inter-X++ focuses on body and hand motions captured in an indoor MoCap environment and does not include facial behavior, as non-professional performers may not consistently produce expressions that match the enacted interactions. Future data collection with professional actors or in more natural settings could better capture the relationship between facial emotion and interactive motion. 2) \textbf{Interaction duration.} The current dataset consists primarily of atomic interactions rather than long, compositional sequences with frequent behavioral transitions. Extending it to continuous scenarios would support the study of more complex social dynamics. 3) \textbf{Hierarchical text-motion alignment.} Our experiments across different description levels provide only an initial investigation of how linguistic granularity affects motion generation; more comprehensive alignment objectives, evaluation protocols, and conditioning strategies remain to be explored. 4) \textbf{Physical interaction.} Although we provide and evaluate physics-based HHI imitation, applying these findings to embodied settings, particularly human-robot interaction, requires further study of contact dynamics, safety constraints, and real-world deployment.

\bibliographystyle{IEEEtran}
\bibliography{egbib}

\vfill

\twocolumn[
\begin{center}
\Large{\bf Inter-X++: A Comprehensive Benchmark for Multimodal Human-Human Interaction Analysis \\ Supplementary Material}
\end{center}
]

\section{Dataset and Annotation Details}

All participants signed informed consent forms authorizing data collection and academic use of the dataset. The released records are de-identified and contain no direct identifiers. To construct the definition of action categories of Inter-X++, we first manually collected candidate categories by referring to established human-human interaction datasets~\cite{chi3d,nturgbd120,interhuman} and then leveraged large language models~\cite{gpt3} to assist in expanding and consolidating the candidate set. The candidates were subsequently subjected to manual screening, semantic deduplication, and feasibility verification in the motion-capture environment. This process resulted in the 40 common daily human-human interaction categories listed in~\cref{tab:actions}. We focus specifically on direct interactions between two people and exclude object-mediated interactions in which the participants interact primarily through a manipulated object.

For the textual annotations, we developed an annotation tool based on~\cite{ait-viewer}, allowing the rendered interaction sequences to be inspected from freely adjustable viewpoints. Each interaction sequence was independently annotated by three annotators, who were instructed to describe the coarse full-body movements, fine-grained hand and finger articulations, relative spatial configurations, and participant-specific behaviors. GPT-3.5~\cite{gpt3} was subsequently used only to refine linguistic expression and correct typographical errors. We then conducted extensive manual screening to verify motion-text consistency, participant roles, temporal order, and the accuracy of fine-grained body-part descriptions; ambiguous or low-quality annotations were revised or removed.

\begin{table*}[t]
  \centering
  \caption{The 40 common daily human-human interaction categories included in the Inter-X++ dataset.}
  \label{tab:actions}
  \begin{tabular}{@{}|l|l|l|l|@{}}
    \toprule
    A01: Hug & A02: Handshake & A03: Wave & A04: Grab \\
    \midrule
    A05: Hit & A06: Kick & A07: Posing & A08: Push \\
    \midrule
    A09: Pull & A10: Sit on leg & A11: Slap & A12: Pat on back \\
    \midrule
    A13: Point finger at & A14: Walk towards & A15: Knock over & A16: Step on foot \\
    \midrule
    A17: High-five & A18: Chase & A19: Whisper in ear & A20: Support with hand \\
    \midrule
    A21: Finger-guessing & A22: Dance & A23: Link arms & A24: Shoulder to shoulder \\
    \midrule
    A25: Bend & A26: Carry on back & A27: Massage shoulder & A28: Massage leg \\
    \midrule
    A29: Hand wrestling & A30: Chat & A31: Pat on cheek & A32: Thumb up \\
    \midrule
    A33: Touch head & A34: Imitate & A35: Kiss on cheek & A36: Help up \\
    \midrule
    A37: Cover mouth & A38: Look back & A39: Block & A40: Fly kiss \\
    \bottomrule
  \end{tabular}
\end{table*}

\section{Broader Impacts}
Inter-X++ can facilitate both generative and perceptive HHI research, with potential applications in digital-human creation, AR/VR, interactive gaming, animation, and human-robot interaction. Its multimodal annotations and standardized evaluation protocols may also improve the reproducibility and comparability of HHI methods. Meanwhile, motion patterns, personality traits, and interpersonal relationships may contain sensitive information, and related models could be misused for intrusive surveillance or behavioral profiling. Since the dataset is collected in a controlled environment and may contain demographic or cultural biases, downstream users should follow appropriate privacy and licensing requirements, evaluate generalization carefully, and conduct additional safety validation before deployment in real-world or embodied systems.

\section{SMPL-X Optimization Details}

We represent each $N$-frame motion sequence using the SMPL-X~\cite{smplx} parametric model. Specifically, the sequence is parameterized by body pose parameters $\theta \in \mathbb{R}^{N\times55\times 3}$, global translations $t\in\mathbb{R}^{N\times3}$, and shape parameters $\beta\in\mathbb{R}^{N\times10}$. We initialize the shape parameters of each subject from the corresponding height and weight, following~\cite{virtual_caliper}. To accurately recover whole-body motion while preserving fine-grained hand articulation, we then fit SMPL-X to the captured MoCap skeletons using a two-stage optimization procedure.

In the first stage, we optimize the body pose parameters while excluding the finger poses. The joint fitting term aligns the SMPL-X joints with the corresponding joints of the captured skeleton:
\begin{equation}
    \mathbb{E}_j=\frac{1}{N}\sum_{i=0}^{N-1}
     \sum_{j\in\mathcal{J}}
     \left\|\bm{J}_j^i\bigl(\mathbb{M}(\theta_b,t)\bigr)-\bm{g}_j^i\right\|_2^2,
\end{equation}
where $\mathcal{J}$ denotes the joint set, $\mathbb{M}$ is the SMPL-X parametric model, $\bm{J}_j^i$ denotes the joint regressor for joint $j$ at frame $i$, $\theta_b$ represents the pose parameters excluding the fingers, and $\bm{g}$ denotes the skeleton captured by the MoCap system. To suppress frame-to-frame jitter and improve temporal coherence, we further introduce the following smoothing term:
\begin{equation}
    \mathbb{E}_{smooth}=\frac{1}{N-1}\sum\limits_{i=0}\limits^{N-1}\sum\limits_{j\in\mathcal{J}}\|\bm{J}_j^{i+1}-\bm{J}_j^{i}\|_2^2
\end{equation}
In addition, we apply a pose regularization term to discourage the optimized body poses from deviating excessively:
\begin{equation}
    \mathbb{E}_{r}=\|\theta_b\|_2^2
\end{equation}
The complete objective for the first stage is therefore defined as:
\begin{equation}
    \mathbb{E}_1=\lambda_j\mathbb{E}_j+\lambda_{smooth}\mathbb{E}_{smooth}+\lambda_r\mathbb{E}_r,
\end{equation}
where we set $\lambda_j,\lambda_{smooth},\lambda_r=1,0.1,0.01$.

In the second stage, we introduce the finger pose parameters and jointly refine the complete whole-body pose. Because accurate hand articulation is particularly important for close human-human interactions, we separate the hand-related terms from the remaining body terms and assign them different optimization weights. The second-stage objective is formulated as:
\begin{gather}
    \mathbb{E}_b=
    \lambda_j\mathbb{E}_j+\lambda_{smooth}\mathbb{E}_{smooth}+\lambda_r\mathbb{E}_r, \\
    \mathbb{E}_{h}=\lambda_{j_{h}}\mathbb{E}_{j_{h}}+\lambda_{{smooth}_h}\mathbb{E}_{{smooth}_h}+\lambda_{r_h}\mathbb{E}_{r_h}, \\
    \mathbb{E}_2=\mathbb{E}_b+\mathbb{E}_h,
\end{gather}
where $\mathbb{E}_b$ and $\mathbb{E}_h$ denote the body and hand objectives, respectively, and $\mathbb{E}_2$ is their combination. We set $\lambda_j,\lambda_{smooth},\lambda_r=1,0.1,0.01$ for the body part and $\lambda_{j_h},\lambda_{{smooth}_h},\lambda_{r_h}=10,0.01,0.001$ for fingers.

\section{Details of Evaluation Metrics}

\subsection{Metrics for Interaction Generation}

Following prior human motion generation benchmarks~\cite{action2motion,actor,mdm,humanml3d,interhuman}, we evaluate generated interactions in task-specific feature spaces. For text-conditioned generation, a motion encoder and a text encoder are trained with a contrastive objective to map paired motions and descriptions into a shared latent space. For action-conditioned interaction generation, human reaction generation, and stylized interaction generation, we use a pretrained ST-GCN~\cite{stgcn} evaluator to extract motion features and predict action labels.

\noindent\textbf{Fr\'echet Inception Distance.}
FID~\cite{fid} measures the distributional discrepancy between real and generated interaction features. A lower FID indicates that the generated interactions more closely match the real-data distribution.

\noindent\textbf{R-Precision.}
R-Precision evaluates motion-text alignment. For each generated interaction, its ground-truth description is mixed with 31 randomly selected mismatched descriptions. All candidate descriptions are ranked according to their distances from the generated motion in the shared embedding space. Top-$k$ R-Precision reports the probability that the ground-truth description appears among the top $k$ retrieved candidates, where we report Top-1, Top-2, and Top-3 results. Higher values indicate better semantic consistency between the generated interaction and the text condition.

\noindent\textbf{MultiModal Distance (MM Dist).}
MM Dist is the average Euclidean distance between the embeddings of generated interactions and their paired textual descriptions. It directly measures cross-modal alignment in the shared motion-text space; a lower value indicates that the generated motions better match the input descriptions.

\noindent\textbf{Action Recognition Accuracy (Acc.).}
For action-conditioned generation tasks, we use the pretrained ST-GCN evaluator to predict the action category of each generated interaction. Accuracy measures the proportion of predictions that match the conditioning action labels, with a higher value indicating better action consistency.

\noindent\textbf{Diversity.}
Diversity measures the overall variation of generated interactions across all conditions. Given two independently sampled sets of motion features $\{v_i\}_{i=1}^{S_d}$ and $\{v_i^\prime\}_{i=1}^{S_d}$, it is computed as
\begin{equation}
    \mathrm{Diversity}=\frac{1}{S_d}\sum_{i=1}^{S_d}
    \|v_i-v_i^\prime\|_2.
\end{equation}
We compute Diversity for the real motions in the same manner and use the resulting value as a reference. For results marked with $\rightarrow$, better performance is indicated by a generated-motion Diversity value closer to the real-motion reference.

\noindent\textbf{Multimodality (MModality).}
MModality quantifies the intra-condition variability of generated interactions. Given $C$ conditions, we independently sample two sets of $S_l$ motions under each condition and compute
\begin{equation}
    \mathrm{MModality}=\frac{1}{C S_l}
    \sum_{c=1}^{C}\sum_{i=1}^{S_l}
    \|v_{c,i}-v_{c,i}^\prime\|_2.
\end{equation}
Here, $v_{c,i}$ and $v_{c,i}^\prime$ denote the feature representations of the paired motion samples generated under condition $c$. A higher MModality value indicates greater variability among interactions generated from the same condition. For results marked with $\rightarrow$, performance is assessed relative to the corresponding real-motion reference, with closer values indicating better agreement.

\subsection{Metrics for Interaction Imitation}

Following PHC~\cite{PHC}, we evaluate physics-based interaction imitation using one success metric and four tracking-error metrics. \textbf{Success rate (Succ)} is the percentage of reference sequences that the simulated humanoids track to completion without triggering the standard failure or early-termination criterion; a higher value is better. \textbf{Global MPJPE ($E_{\mathrm{g\text{-}mpjpe}}$)} measures the mean Euclidean distance between reference and simulated joint positions in the global coordinate system and therefore includes global translation errors. \textbf{MPJPE ($E_{\mathrm{mpjpe}}$)} computes the joint-position error after removing the global root translation, focusing on relative pose accuracy. \textbf{Acceleration error ($E_{\mathrm{acc}}$)} and \textbf{velocity error ($E_{\mathrm{vel}}$)} measure the average discrepancies between the reference and simulated joint accelerations and velocities, respectively. Lower values for all four error metrics indicate more accurate and temporally coherent imitation.

\subsection{Metrics for Perceptive Tasks}

\noindent\textbf{Interaction Recognition and Causal-Order Accuracy.}
For HHI recognition, Top-1 and Top-5 accuracy measure whether the ground-truth interaction category is the highest-scoring prediction or is included among the five highest-scoring predictions, respectively. For causal interaction order inference, binary accuracy is the proportion of sequences for which the actor-reactor order is correctly identified. Higher values indicate better recognition performance.

\noindent\textbf{Captioning Metrics.}
We evaluate motion-to-text interaction captioning using complementary lexical and semantic metrics. \textbf{BLEU@1} and \textbf{BLEU@4}~\cite{bleu} measure unigram and up-to-four-gram precision, together with a brevity penalty, between generated and reference descriptions. \textbf{ROUGE}~\cite{rouge} emphasizes recall-oriented lexical overlap and measures how much salient reference content is preserved. \textbf{CIDEr}~\cite{cider} computes consensus agreement using TF--IDF-weighted $n$-grams across the reference captions. \textbf{BERTScore}~\cite{bertscore} matches contextualized token embeddings and is therefore more sensitive to semantic similarity than exact word overlap. Higher values are preferred for all captioning metrics.

\noindent\textbf{Personality Assessment.}
We report the coefficient of determination ($R^2$) independently for Openness, Conscientiousness, Extraversion, Agreeableness, and Neuroticism. Given ground-truth scores $y_i$, predictions $\hat{y}_i$, and the ground-truth mean $\bar{y}$, it is defined as
\begin{equation}
    R^2=1-\frac{\sum_i(y_i-\hat{y}_i)^2}
    {\sum_i(y_i-\bar{y})^2}.
\end{equation}
An $R^2$ of 1 indicates perfect prediction, 0 corresponds to predicting the mean, and a negative value indicates performance worse than the mean predictor.

\subsection{Repeated Evaluation and Result Interpretation}

Unless otherwise specified, stochastic generation evaluations are repeated 20 times with different random seeds, and we report the mean together with the 95\% confidence interval. MModality for text-conditioned generation is evaluated five times because each evaluation requires multiple samples for every text condition. In the result tables, $\uparrow$ and $\downarrow$ indicate that higher and lower values are preferred, respectively, whereas $\rightarrow$ indicates that proximity to the real-data statistic is preferred.

\section{Additional Implementation Details}
\label{sec:supp_impl}

\paragraph{Data representation and preprocessing}
We use the official Inter-X++ training, validation, and test splits, with all motion sequences represented at 30~fps. Each two-person interaction of length $T$ is stored as a tensor in $\mathbb{R}^{T\times56\times12}$. After separating the two actors, each frame is represented by a $56\times6$ feature matrix consisting of the continuous 6D rotations of 55 SMPL-X joints and one root-trajectory entry. The global coordinate system is retained to preserve the relative spatial configuration between the participants.

For Stage~1, training samples are extracted as 64-frame windows with a temporal stride of 10 frames. For Stage~2, sequences shorter than 24 frames are excluded, motion lengths are quantized in units of four frames, and sequences are cropped and padded or truncated to a maximum length of 150 frames. The original Inter-X++ captions are used, with a maximum length of 35 tokens.

\subsection{Stage 1: Caption-Aware Unified Motion Tokenizer}
\label{sec:supp_vq}

\paragraph{Tokenizer architecture}
The first stage employs a two-dimensional convolutional residual VQ-VAE with parameters shared between the two actors. Two strided encoder blocks reduce the temporal resolution by a factor of four and aggregate the 56 skeletal entries into five spatial groups. The encoder--decoder width and code dimension are set to 512. Vector quantization uses a codebook of 1,024 embeddings updated by an exponential moving average with a decay rate of $0.99$.

We introduce two-layer pre-normalization Transformers immediately before and after vector quantization. Both modules use 512-dimensional features, eight attention heads, a feed-forward network width of 2,048, GELU activations, and a dropout rate of $0.1$. Their residual branches are zero-initialized to preserve the initial behavior of the VQ-VAE.

\paragraph{Caption-aware objective}
The latent features of the two actors are concatenated and mean-pooled to condition a single-layer GRU caption decoder with a hidden dimension of 512. The decoder is trained with teacher forcing and token-level cross entropy. The complete Stage~1 objective is
\begin{equation}
\begin{aligned}
\mathcal{L}_{\mathrm{VQ}} ={}&
\mathcal{L}_{\mathrm{rec}}
+0.02\mathcal{L}_{\mathrm{commit}}
+\mathcal{L}_{\mathrm{pos}} \\
&+100\mathcal{L}_{\mathrm{vel}}
+5\mathcal{L}_{\mathrm{bone}}
+0.01\mathcal{L}_{\mathrm{geo}} \\
&+500\mathcal{L}_{\mathrm{foot}}
+0.5\mathcal{L}_{\mathrm{cap}}.
\end{aligned}
\end{equation}
Here, the reconstruction and commitment terms supervise motion tokenization; the position, velocity, bone-length, geodesic, and foot-contact terms enforce kinematic consistency; and $\mathcal{L}_{\mathrm{cap}}$ provides interaction-level semantic supervision.

\paragraph{Optimization}
The tokenizer is optimized using AdamW with a learning rate of $2\times10^{-4}$, $\beta=(0.9,0.99)$, and zero weight decay. Gradients are clipped to a maximum norm of 1.0. Training is conducted for 50 epochs with a batch size of 128.

\subsection{Stage 2: Text-Conditioned Masked Interaction Transformer}
\label{sec:supp_transformer}

Our text-conditioned interaction generator builds upon InterMask~\cite{javed2024intermask} and is adapted to the unified motion tokenizer. The Stage~1 tokenizer remains fixed during this stage. The two actor-specific token sequences are concatenated using a learned separation token. The vocabulary consists of 1,024 motion codes and three special symbols for masking, padding, and actor separation. Each 512-dimensional code embedding is projected into a 384-dimensional Transformer space and augmented with spatial and temporal positional encodings.

\paragraph{Interaction Transformer and text conditioning}
The masked interaction Transformer comprises six blocks, each with six attention heads, a feed-forward network width of 1,024, and a dropout rate of $0.2$. The two actors are modeled jointly through self-attention and cross-actor attention. Textual conditions are encoded using a frozen CLIP ViT-L/14@336px backbone. A trainable two-layer text Transformer with a width of 768, eight attention heads, and a feed-forward network width of 2,048 projects the text representation into the motion-Transformer space. The text condition is dropped with a probability of $0.1$ during training to enable classifier-free guidance.

\paragraph{Training}
The masking ratio follows a cosine schedule, and cross-entropy loss is evaluated at the selected token positions. Interaction masking is applied with a probability of $0.2$ to encourage the model to infer one participant from the motion of the other. The masked interaction Transformer is trained from random initialization using AdamW with a learning rate of $2\times10^{-4}$, $\beta=(0.9,0.99)$, and a weight decay of $10^{-5}$. Training is conducted for 500 epochs with a batch size of 128.

\subsection{Inference and Evaluation}
\label{sec:supp_eval}

\paragraph{Sampling}
Generation begins from fully masked two-actor token sequences. Tokens are iteratively sampled, and low-confidence predictions are re-masked according to a cosine schedule. Unless otherwise specified, we use 20 refinement steps, categorical sampling with a temperature of 1.0, and a classifier-free guidance scale of 2. The final actor-specific token sequences are decoded independently by the shared VQ decoder.

\paragraph{Evaluation protocol}
We evaluate on the Inter-X++ test split with a batch size of 32 using an independently trained and frozen text--motion evaluator. Text inputs are represented using 300-dimensional GloVe embeddings and part-of-speech features, while text and motion are mapped into a shared 512-dimensional embedding space. We report Matching Score, R-Precision at ranks 1--3, FID, Diversity, and Multimodality. Following the protocol described above, stochastic evaluations are repeated 20 times and reported with 95\% confidence intervals, whereas Multimodality is evaluated five times. Diversity is computed using 300 randomly sampled motion pairs. For Multimodality, we sample 100 captions, generate 30 motions for each caption, and perform 10 random pairwise comparisons.

\paragraph{Software}
The implementation uses Python 3.9.13, PyTorch 1.13.1, and CUDA 11.7. Training is performed using a single-process, single-GPU pipeline.

\section{Additional Qualitative Results on Inter-X++}
\begin{figure*}[t]
  \centering
  \includegraphics[width=1.0\linewidth]{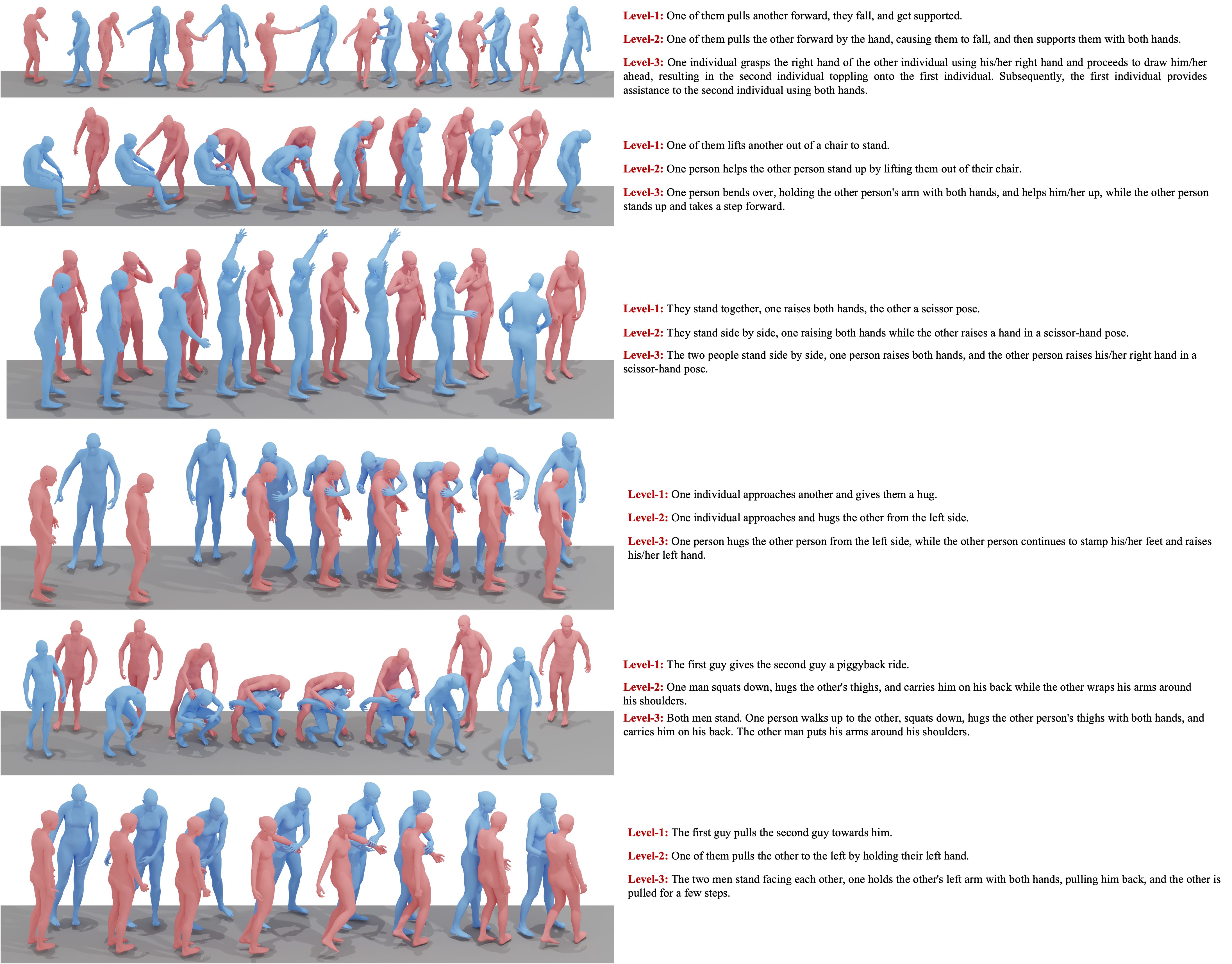}
  \caption{Examples of the hierarchical textual annotations in Inter-X++. For each interaction, Level-1 gives a concise action summary, Level-2 describes the overall interaction in greater detail, and Level-3 additionally specifies participant-specific movements, body parts, and spatial relations.}
  \label{fig:suppl_hierarchical_texts}
\end{figure*}

\begin{figure*}[t]
  \centering
  \includegraphics[width=1.0\linewidth]{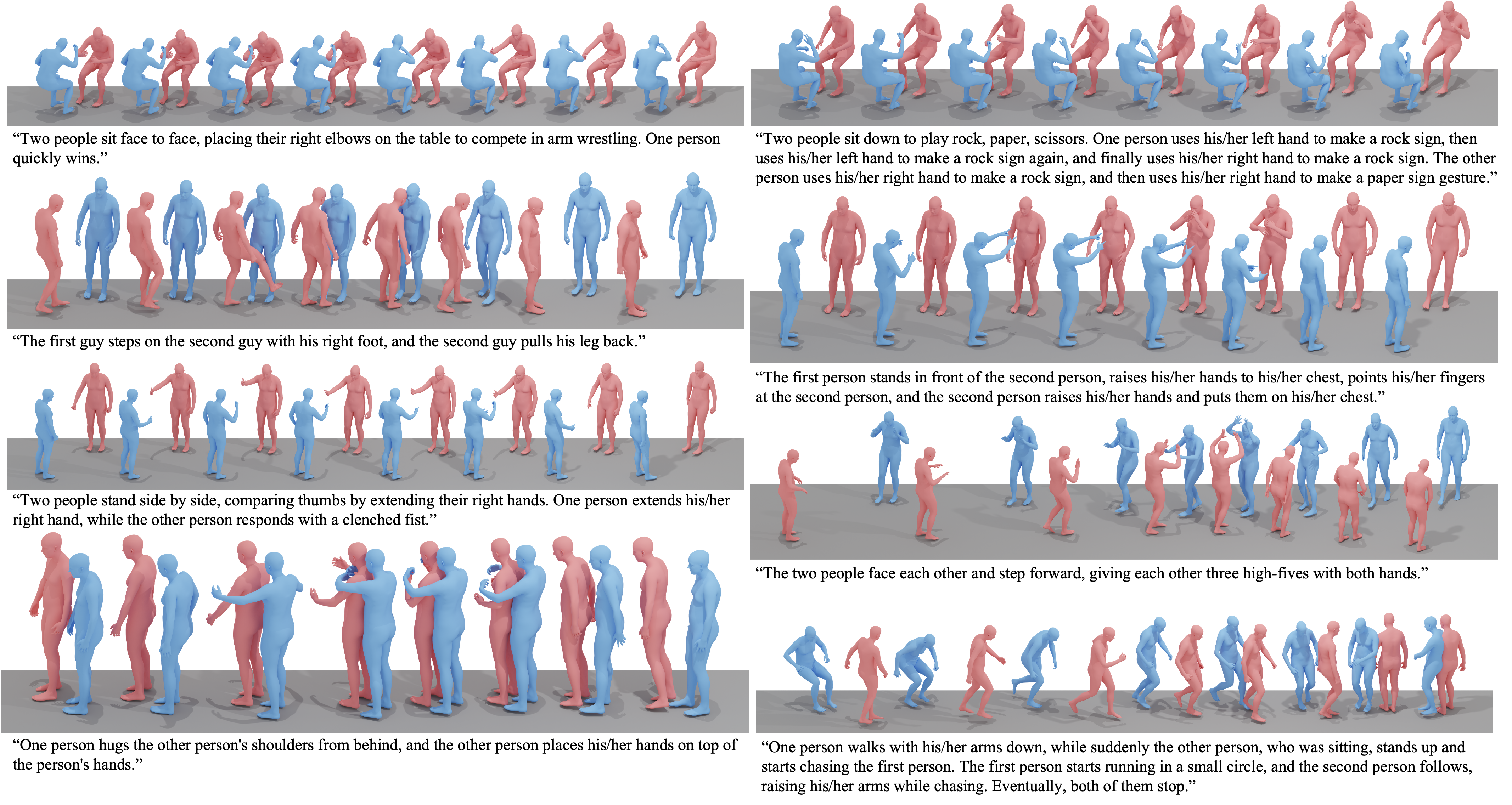}
  \caption{Additional qualitative results of text-conditioned human-human interaction generation on Inter-X++. Each example shows a generated interaction sequence together with its input description.}
  \label{fig:suppl_hhi_gen}
\end{figure*}

\begin{figure*}[t]
  \centering
  \includegraphics[width=1.0\linewidth]{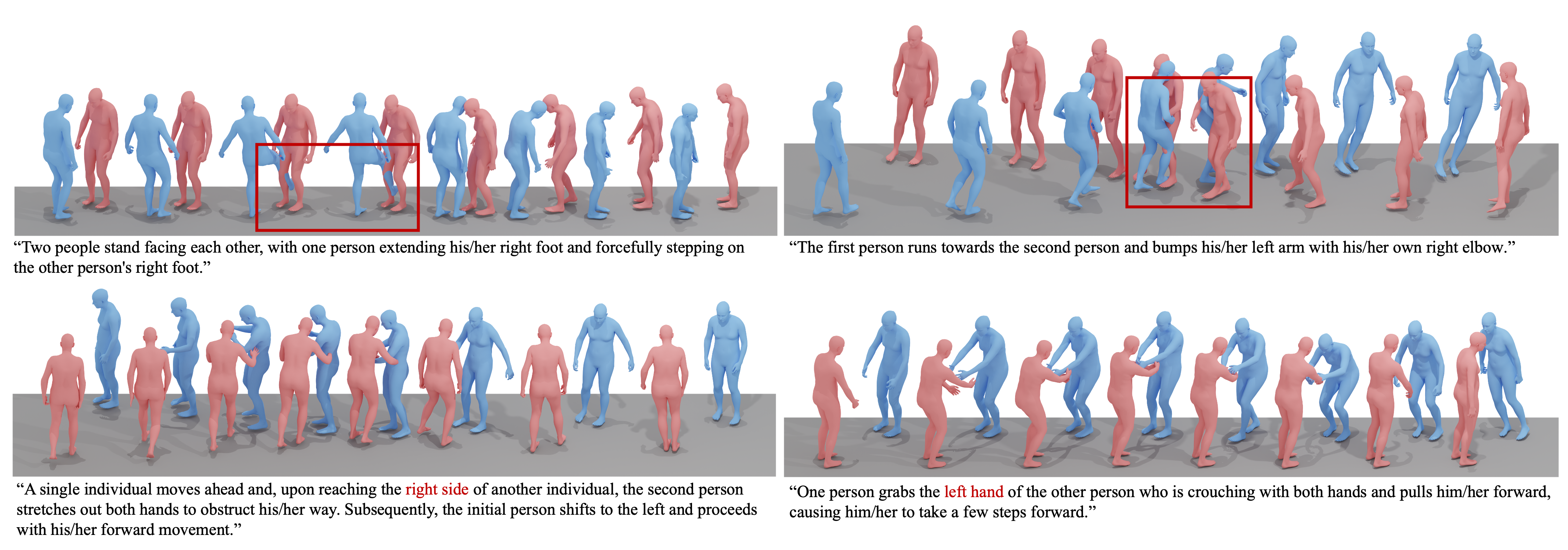}
  \caption{Representative failure cases of text-conditioned human-human interaction generation. The top row highlights frames with inaccurate or physically implausible contacts using red boxes, while the bottom row illustrates inaccurate grounding of spatial and body-part constraints, which are emphasized in red in the input descriptions.}
  \label{fig:suppl_hhi_gen_fail}
\end{figure*}

\begin{figure*}[t]
  \centering
  \includegraphics[width=1.0\linewidth]{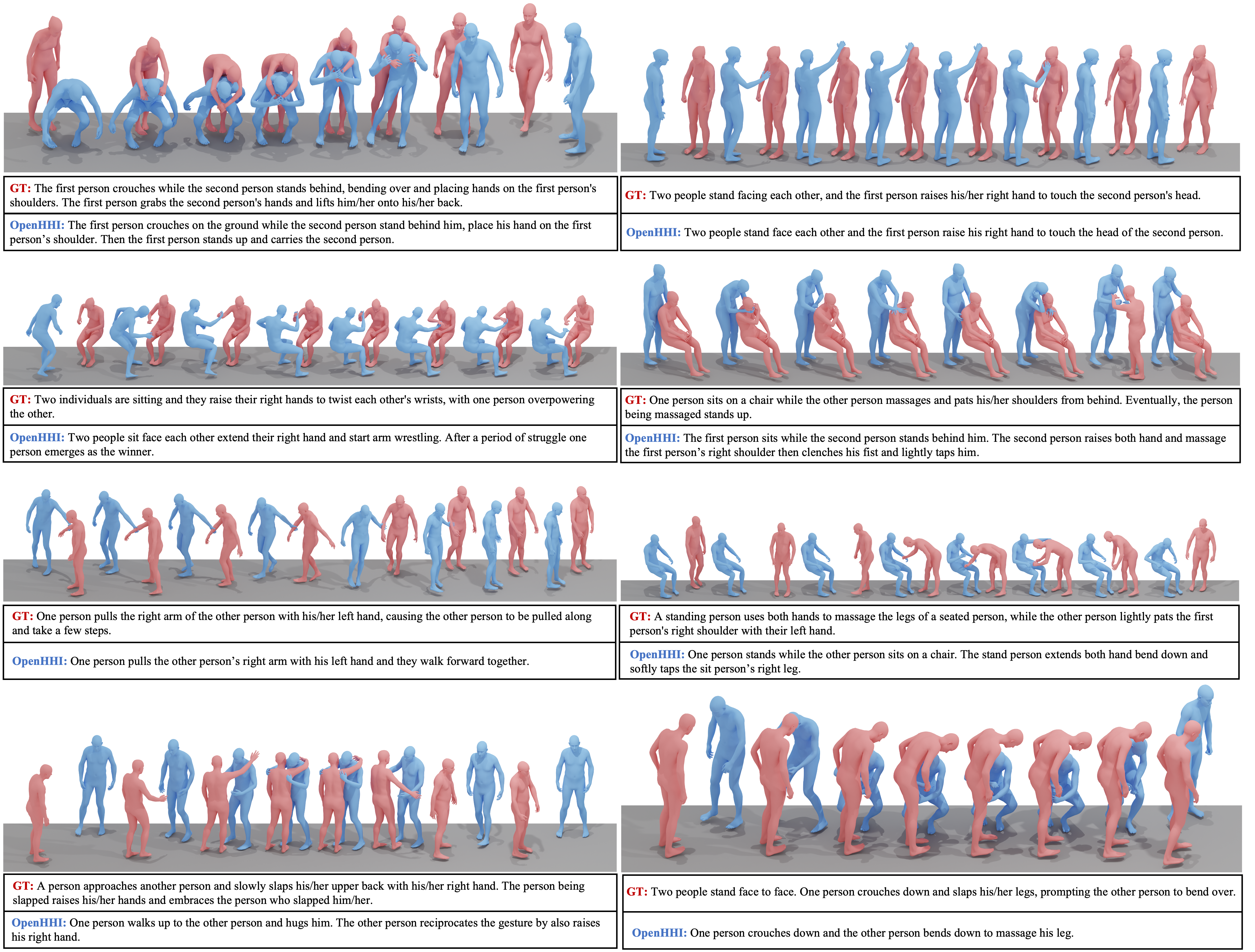}
  \caption{Additional qualitative results of human-human interaction captioning on Inter-X++. Each example presents a sampled interaction sequence together with its ground-truth description and the caption generated by OpenHHI.}
  \label{fig:suppl_captioning}
\end{figure*}

This section presents additional qualitative results for the hierarchical textual annotations, text-conditioned human-human interaction generation, and motion-to-text interaction captioning on Inter-X++. As illustrated in~\cref{fig:suppl_hierarchical_texts}, the three annotation levels exhibit progressively increasing semantic granularity, ranging from concise action summaries to detailed descriptions of participant-specific movements, body-part dynamics, and spatial relations. The examples in~\cref{fig:suppl_hhi_gen} demonstrate how textual conditions govern the actions, relative configurations, and coordinated motions of the two participants, whereas~\cref{fig:suppl_hhi_gen_fail} highlights representative failure modes. Moreover,~\cref{fig:suppl_captioning} compares the captions predicted by OpenHHI with the corresponding ground-truth descriptions, providing a qualitative assessment of motion--language semantic correspondence.

\end{document}